\documentclass[10pt]{article}
\usepackage{leapwing}
\usepackage{etoolbox}

\usepackage{amsmath,amsfonts,bm}

\def\eqref#1{equation~\ref{#1}}

\def\1{\bm{1}}

\DeclareMathAlphabet{\mathsfit}{\encodingdefault}{\sfdefault}{m}{sl}
\SetMathAlphabet{\mathsfit}{bold}{\encodingdefault}{\sfdefault}{bx}{n}

\newif\ifshowchanges
\showchangesfalse       

\DeclareRobustCommand{\rev}[1]{%
    \ifshowchanges
        {\color{blue}#1}%
    \else
        #1%
    \fi
}

\renewcommand{\LWTitle}{SOAR: temporal Structure-guided One-step Action generation for Robotic manipulation}
\renewcommand{\LWShortTitle}{SOAR}
\renewcommand{\LWDate}{November 2026}
\renewcommand{\LWPartnerLogoHeight}{11mm}
\renewcommand{\LWPartnerLogos}{%
  \LWPartnerLogo[12mm]{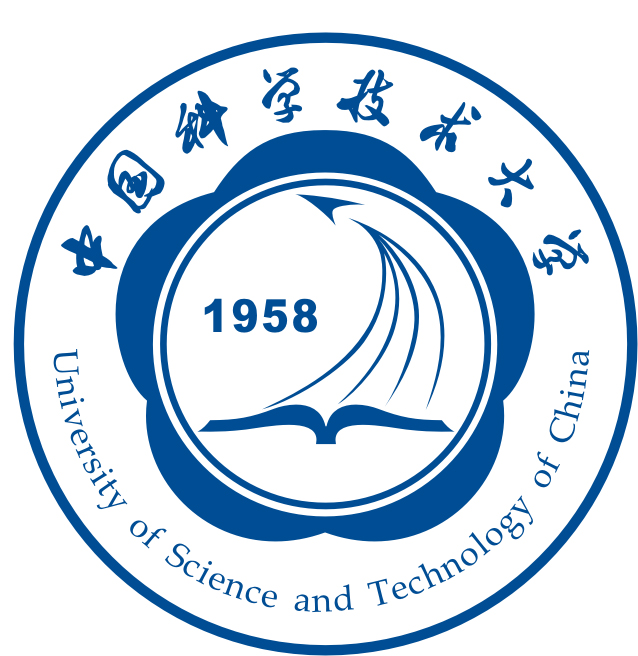}
}
\renewcommand{\LWAuthors}{%
Di Xin\textsuperscript{1*},
Mingfei Han\textsuperscript{1,2*\textdagger},
Genming Cui\textsuperscript{1},
Yuanxiang Chen\textsuperscript{1},
Xiaolong Wu\textsuperscript{1},
Siyu Ning\textsuperscript{1},
Xiaojun Chang\textsuperscript{1,2},
Zhihui Li\textsuperscript{1}
}
\renewcommand{\LWAffiliations}{%
    \textsuperscript{1} Future Media Computing Lab, University of Science and Technology of China~
    \textsuperscript{2} LeapWing AI
}
\renewcommand{\LWNotes}{%
  {\sffamily\color{LWMuted}
  \textsuperscript{*}Equal contribution\quad
  \textsuperscript{\textdagger}Project lead\quad
    }
}

\renewcommand{\LWAbstract}{%
\rev{Efficient robotic manipulation requires temporally
coordinated actions with low inference latency.
Multi-step diffusion and flow policies progressively
refine action trajectories but require costly iterative
inference, whereas one-step policies eliminate iterative
trajectory refinement to reduce inference cost.}
We introduce \textbf{SOAR}, a structure-guided one-step generative policy that improves action quality while preserving one-step efficiency. 
\rev{SOAR predicts an observation-conditioned discrete cosine transform (DCT) representation of future actions, capturing temporal variations across multiple frequency bands to provide explicit guidance from key temporal structural information.}
Unlike velocity-parameterized one-step flow policies that recover clean actions indirectly from a predicted velocity field, SOAR directly generates clean action trajectories in action space. 
This aligns the model output with the executable control target and provides a direct interface for structural calibration. The predicted structure further corrects the generated trajectory according to their spectral discrepancy, without requiring an additional evaluation of the generative backbone. 
During training, the corresponding MeanFlow velocity is derived from the calibrated clean action and used for consistency learning, while an endpoint objective aligns training with the pure-noise condition used for one-step inference. 
Across 53 Adroit and Meta-World tasks, SOAR achieves 83.9\% overall success with only 8.7~ms inference latency. When applied to the VLA model, \textit{$\pi_{0.5}$}, SOAR further achieves 97.2\% on LIBERO and 71.7\% on RoboTwin 2.0. Real-world experiments and extensive ablations further validate its effectiveness and efficiency.
}

\begin{document}
\makeLWTitle

\section{Introduction}

Learning visuomotor policies from expert demonstrations enables robots to acquire complex manipulation skills without manually designing task-specific controllers. 
However, given the same observation and task goal, multiple action trajectories may succeed while differing in approach direction, motion timing, or contact strategy.
Unlike deterministic behavior cloning, which can average across distinct action modes, 
diffusion- and flow-based policies address this ambiguity by modeling the action distribution itself and have shown strong performance in robotic manipulation~\citep{chi2023diffusion,zhang2025flowpolicy}.


Despite their effectiveness, generative policies often rely on iterative inference. Diffusion policies progressively denoise action trajectories, while flow-based policies typically integrate learned velocity fields over multiple steps. 
Such interactive strategy repeats network evaluations, introducing substantial inference latency
that reduces responsiveness to dynamics and can leave predicted actions stale as the scene evolves~\citep{black2025realtimechunking}. 
Some recent efforts seek to reduce inference steps through accelerated sampling, distillation, consistency learning, or shortened flow trajectories~\citep{song2021ddim,salimans2022progressive,prasad2024consistencypolicy,hu2024adaflow,su2025freqpolicy}, even pushing them to a single network evaluation~\citep{geng2025meanflow,sheng2026mp1,fang2026omp}. 
However, such efficiency comes at the cost of iterative refinement. 
Multi-step generation repeatedly revisits the action trajectory, progressively correcting its temporal organization; a one-step policy must instead determine the entire trajectory structure at once, with no subsequent opportunity for correction.


\begin{figure}[t]
    \centering
    \includegraphics[width=\textwidth]{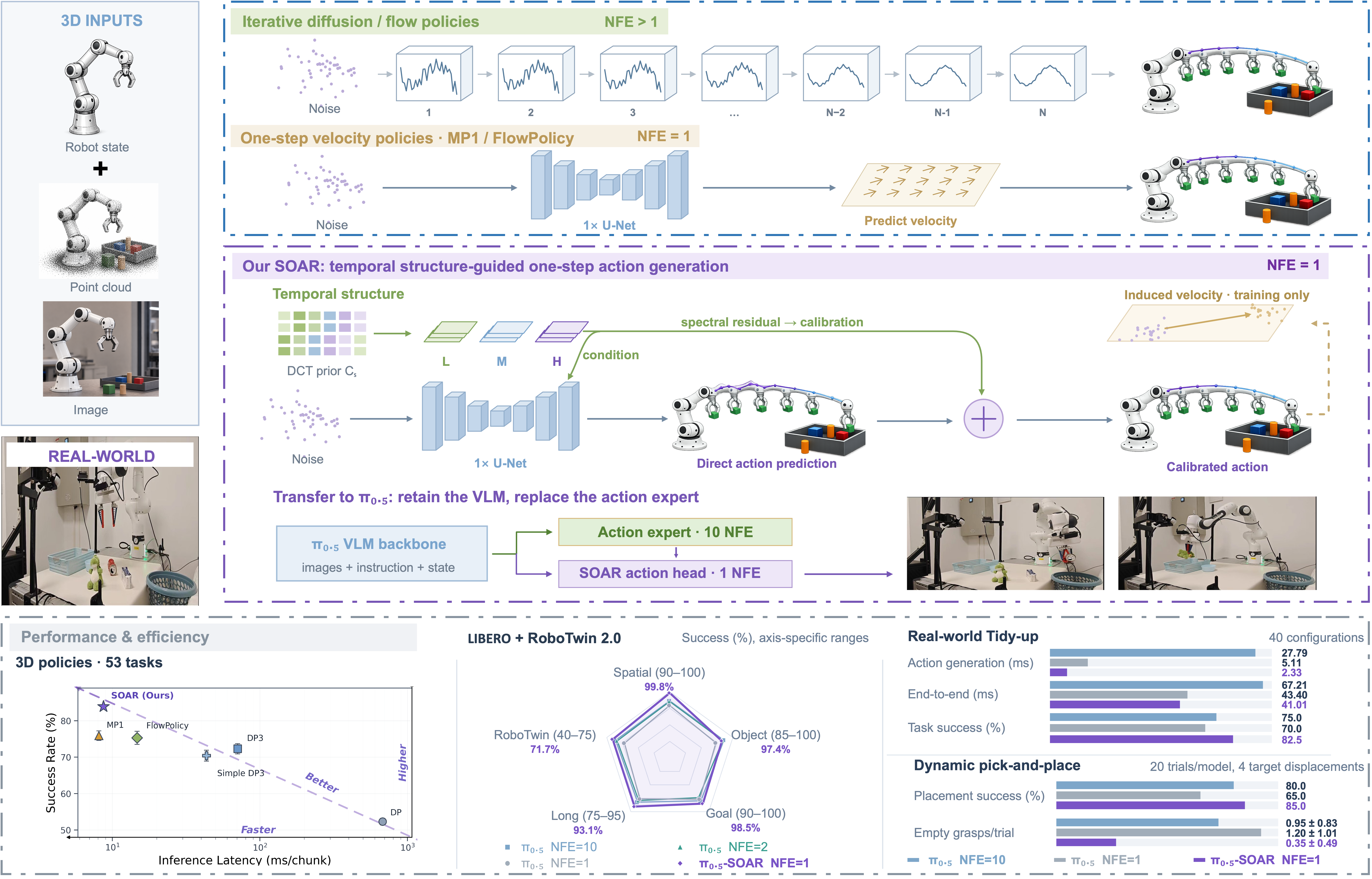}
    \caption{
    \textbf{Teaser of SOAR.}
    Iterative diffusion and flow policies require multiple refinement steps, while existing one-step policies typically generate actions through a learned velocity field.
    SOAR instead directly predicts a clean action chunk with a single U-Net evaluation, guided and calibrated by an observation-conditioned frequency representation; the corresponding transport velocity is used only for training.
    The same design replaces the multi-step action expert of $\pi_{0.5}$ while retaining its VLM backbone.
    Across Adroit~\citep{rajeswaran2018dexterous}, Meta-World~\citep{yu2020metaworld}, LIBERO~\citep{liu2023libero}, RoboTwin2.0~\cite{chen2025robotwin}, and real-world steps, SOAR achieves competitive success rates with substantially lower action-generation latency at $\mathrm{NFE}=1$.
    }
    \label{fig:teaser}
\end{figure}

\rev{To provide explicit temporal structure guidance
without relying on iterative refinement, we propose
\textbf{SOAR}, a temporal \textbf{S}tructure-guided
\textbf{O}ne-step \textbf{A}ction generation framework
for \textbf{R}obot manipulation, as shown in
Fig.~\ref{fig:teaser}.
Unlike velocity-parameterized one-step flow policies
that recover clean actions indirectly, SOAR directly
generates clean action trajectories in action space,
aligning the model output with the executable control
target and enabling direct structural calibration.
Given the current observation, SOAR predicts a
multi-band discrete cosine transform (DCT)
representation of future actions to provide temporal
structure guidance~\citep{ahmed1974dct,zhong2025freqautoregressive}.
The predicted structure is incorporated into the
observation condition to guide base action generation
and is subsequently used to calibrate the generated
trajectory according to their spectral discrepancy,
without additional evaluation of the generative
backbone. During training, the calibrated action is mapped to
the corresponding MeanFlow velocity for consistency
and endpoint supervision.
SOAR thus improves action quality while retaining
$\mathrm{NFE}=1$ inference and millisecond-scale latency.
}

Our contributions are:
\begin{itemize}

\item We propose \textbf{SOAR}, a direct one-step action generation policy that generates clean action trajectories in action space with $\mathrm{NFE}=1$ and millisecond-scale inference.

\rev{\item We introduce temporal structure-guided generation and calibration, which explicitly models key temporal structural characteristics of action trajectories and corrects structural discrepancies in generated actions without additional backbone evaluations.}

\item SOAR consistently outperforms compared one-step policies on Adroit and Meta-World while retaining $\mathrm{NFE}=1$, and further demonstrates competitive performance in the VLA setting on LIBERO and RoboTwin2.0.

\end{itemize}

\section{Related Work}

\noindent\textbf{Visual Inputs for Robot Learning.}
Visuomotor policies commonly operate on either image observations or explicit 3D scene representations. Image-based methods such as Behavior Transformer, ACT, and RT-1 use learned visual representations for multimodal action prediction, action chunking, and large-scale robot control~\citep{shafiullah2022bet,zhao2023act,brohan2023rt1}. Because images encode geometry only through projection, several works instead reason explicitly in 3D. PerAct, Act3D, RVT, and GNFactor represent scenes using voxelized features, continuous 3D features, virtual views, or neural feature fields~\citep{gervet2023act3d,goyal2023rvt,ze2023gnfactor}. Explicit 3D representations have also been combined with generative action models, including 3D Diffuser Actor, DP3, and EquiBot~\citep{ke2024diffuseractor,ze2024dp3,yang2025equibot}. These studies highlight the value of 3D geometry for precise and data-efficient manipulation, although observation modality alone does not determine action-generation cost.

\noindent\textbf{Diffusion-based Visuomotor Policies.}
Diffusion policies represent multimodal action distributions through progressive denoising~\citep{ho2020ddpm}. Diffusion Policy and subsequent methods such as ChainedDiffuser and HDP use iterative refinement to generate structured action trajectories~\citep{chi2023diffusion,xian2023chaineddiffuser,ma2024hdp}. Reducing this inference cost has motivated accelerated samplers and distillation. DDIM shortens the sampling trajectory, while progressive distillation compresses multiple denoising steps into fewer evaluations~\citep{song2021ddim,salimans2022progressive}. Consistency Models further enable one- or few-step sampling by enforcing consistency along the generative trajectory~\citep{song2023consistency,song2024improvedconsistency}. In robot learning, Consistency Policy, ManiCM, SDM Policy, and OneDP apply related consistency, distribution-matching, or distillation objectives to accelerate action generation~\citep{prasad2024consistencypolicy,gao2026manicm,jia2024sdm,wang2024onedp}. These methods reduce the cost of iterative denoising, while many one-step variants obtain their one-step mapping by compressing or distilling a multi-step generative process.

\noindent\textbf{Flow-based Visuomotor Policies.}
Flow Matching models generation as transport under a learned time-dependent velocity field~\citep{lipman2023flowmatching}. Standard flow inference numerically integrates this field and therefore requires repeated network evaluations. Rectified Flow reduces integration difficulty by encouraging straighter transport paths, while AdaFlow adaptively determines the number of integration steps~\citep{liu2023rectifiedflow,hu2024adaflow}. Consistency Flow Matching instead imposes consistency along flow trajectories to support efficient generation~\citep{yang2024consistencyfm}. For robotic manipulation, FlowPolicy applies consistency-based flow generation to point-cloud-conditioned control, and FreqPolicy introduces frequency-domain constraints for action-sequence consistency~\citep{zhang2025flowpolicy,su2025freqpolicy}. MeanFlow removes numerical integration at inference by directly predicting interval-averaged velocity~\citep{geng2025meanflow}, providing the basis for native one-step policies such as MP1, OMP, and OFP~\citep{sheng2026mp1,fang2026omp,li2026ofp}. SOAR builds on this one-step formulation but introduces explicit trajectory-structure conditioning and direct action-space correction to improve the generated trajectory without additional sampling iterations.

\section{Method}
\label{sec:method}

\subsection{Overview}
\label{sec:overview}

\begin{figure}[t]
    \centering
    \includegraphics[width=0.98\textwidth]{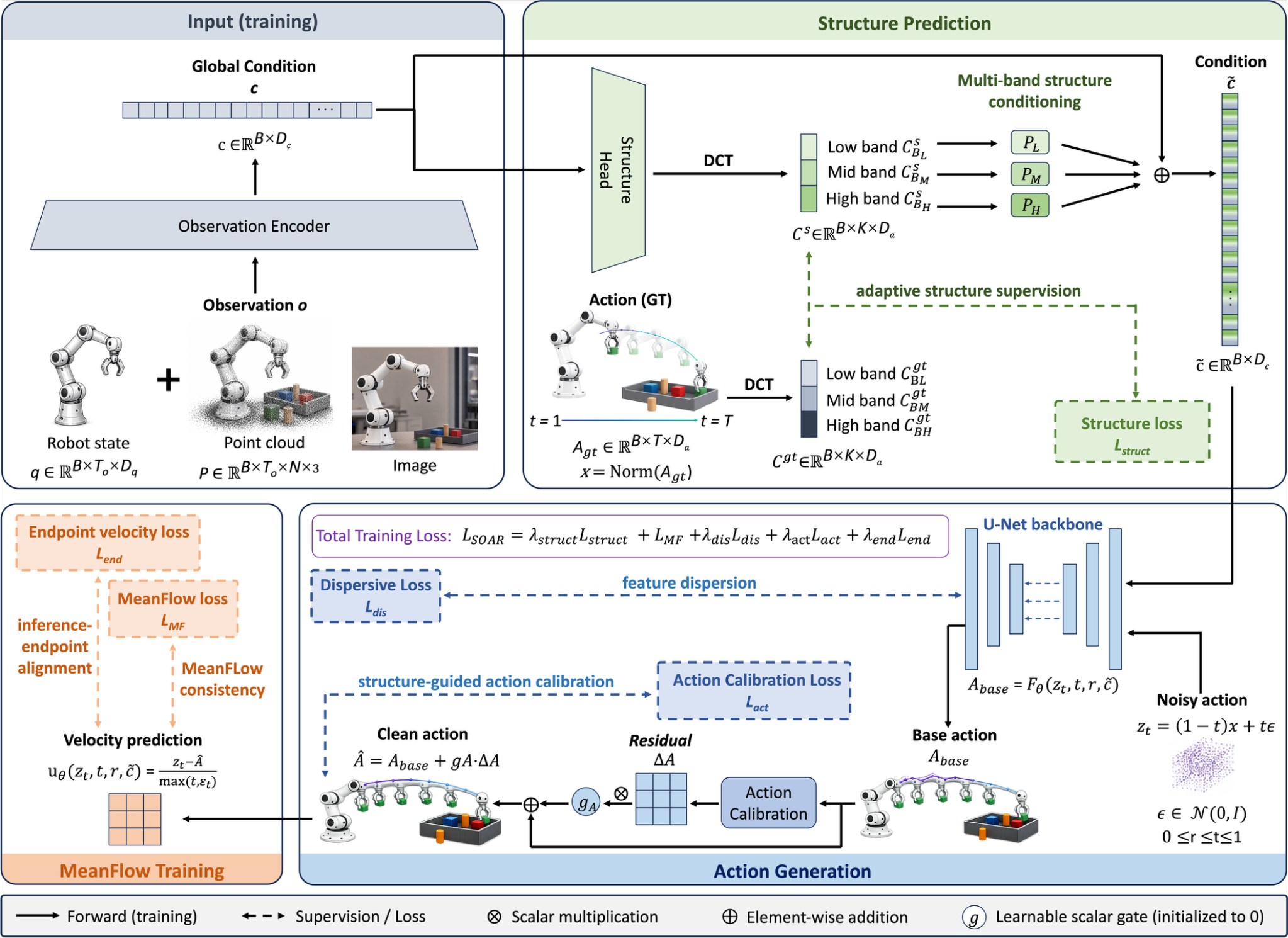}
    \caption{
    Overview of \textbf{SOAR}. Given an observation, SOAR predicts the
    frequency-domain structure of the future action trajectory and injects
    its multi-band features into the action generator. The conditioned U-Net
    directly generates a clean action trajectory, which is further calibrated
    using the predicted structure. During training, the calibrated action
    induces the corresponding MeanFlow transport velocity for consistency
    learning and inference-endpoint alignment. Inference requires a single
    U-Net evaluation.
    }
    \label{fig:soar}
\end{figure}

Given the latest $T_o$ observations
$\mathbf{o}_{1:T_o}
=\{\mathbf{P}_{1:T_o},\mathbf{q}_{1:T_o}\}$,
we denote the normalized demonstrated action chunk as
$\mathbf{x}
=\operatorname{Norm}(\mathbf{A}^{\mathrm{gt}}_{1:T})
\in\mathbb{R}^{B\times T\times D_a}$.
An observation encoder $E_{\phi}$ maps the observations to a global
condition
$\mathbf{c}=E_{\phi}(\mathbf{o}_{1:T_o})$.

As shown in Fig.~\ref{fig:soar}, SOAR directly generates clean action
trajectories in action space rather than recovering them from a predicted
velocity field. An observation-conditioned frequency representation provides
temporal guidance for one-step generation, and the resulting action is further
calibrated according to its spectral discrepancy with the predicted structure.
During training, the calibrated action induces the corresponding MeanFlow
transport velocity for consistency learning and inference-endpoint alignment,
while inference requires only a single evaluation of the action generator.

The overall objective is
\begin{equation}
    \mathcal{L}_{\mathrm{SOAR}}
    =
    \lambda_{\mathrm{struct}}\mathcal{L}_{\mathrm{struct}}
    +
    \lambda_{\mathrm{act}}\mathcal{L}_{\mathrm{act}}
    +
    \lambda_{\mathrm{dis}}\mathcal{L}_{\mathrm{dis}}
    +
    \mathcal{L}_{\mathrm{MF}}
    +
    \lambda_{\mathrm{end}}\mathcal{L}_{\mathrm{end}}.
    \label{eq:total_loss}
\end{equation}
Here, $\mathcal{L}_{\mathrm{struct}}$ supervises trajectory structure,
$\mathcal{L}_{\mathrm{act}}$ constrains the calibrated action,
$\mathcal{L}_{\mathrm{MF}}$ is the standard MeanFlow objective,
$\mathcal{L}_{\mathrm{dis}}$ is the dispersive feature regularizer inherited
from MP1~\citep{sheng2026mp1}, and
$\mathcal{L}_{\mathrm{end}}$ aligns training with the one-step inference
condition.

\subsection{Frequency-Domain Trajectory-Structure Learning}
\label{sec:frequency_space}

\rev{A one-step generator must determine the temporal
organization of an entire action chunk without
iterative refinement. To provide explicit structural
guidance under this constraint, we predict the
temporal structure of future actions in the frequency
domain. Using the discrete cosine transform
(DCT)~\citep{ahmed1974dct},
$\mathbf{C}^{\mathrm{gt}} =
\operatorname{DCT}(\mathbf{x})$ and
$\mathbf{C}^{s} = H_{\psi}(\mathbf{c})$,
where $\mathbf{C}^{\mathrm{gt}}$ is used only for
training supervision and $\mathbf{C}^{s}$ is
predicted from the observation. The structure head serves as an auxiliary
observation-conditioned estimate for guiding action
generation, rather than as a standalone policy
for directly executing predicted trajectories. The residual fusion preserves the original
observation condition while incorporating
complementary temporal structure information
for subsequent action generation.}

\noindent\textbf{Multi-band structure conditioning.}
To capture temporal variations at different scales, we decompose
$\mathbf{C}^{s}$ into low-, mid-, and high-frequency components and project
them into the observation feature space. The resulting multi-band features are
fused with the observation condition to obtain a structure-enhanced
representation $\widetilde{\mathbf{c}}$, which conditions the subsequent
one-step action generator. The frequency partition and fusion mechanism are
described in Appendix~\ref{app:structure_details}.

\noindent\textbf{Adaptive structure supervision.}
DCT coefficients can differ substantially in magnitude, causing a uniform
spectral objective to overemphasize high-energy components. We therefore
weight each coefficient according to its prediction error relative to its
target energy:
\begin{equation}
\vspace{-2mm}
    \mathcal{L}_{\mathrm{struct}}
    =
    \frac{1}{B}
    \sum_{b=1}^{B}
    \sum_{k=0}^{K-1}
    \omega_{b,k}e_{b,k}.
    \label{eq:structure_loss}
\end{equation}
Here, $e_{b,k}$ denotes the spectral prediction error and
$\omega_{b,k}$ its adaptive weight. Frequencies that remain difficult to
predict relative to their own scale receive stronger supervision.
Their exact definitions are given in Appendix~\ref{app:structure_details}.
The predicted spectrum $\mathbf{C}^{s}$ is subsequently used for both action
conditioning and calibration.

\subsection{Structure-Guided Action Generation and Calibration}
\label{sec:action_space}

For training, we follow the standard linear probability path used in
flow-based generative modeling~\citep{lipman2023flowmatching,geng2025meanflow}
to construct an intermediate noisy action state $\mathbf{z}_t$ between the
demonstrated action and Gaussian noise. Details are provided in
Appendix~\ref{app:meanflow_details}.

Conditioned on the structure-enhanced representation
$\widetilde{\mathbf{c}}$, a FiLM-conditioned 1D U-Net
~\citep{perez2018film,chi2023diffusion} directly predicts the clean base
action
\begin{equation}
    \mathbf{A}_{\mathrm{base}}
    =
    F_{\theta}
    (\mathbf{z}_t,t,r,\widetilde{\mathbf{c}}).
    \label{eq:base_action}
\end{equation}
Unlike velocity-parameterized policies that recover the action from a
predicted transport field, $F_{\theta}$ directly outputs the clean action
trajectory in action space.

\noindent\textbf{Structure-guided action calibration.}
Structural conditioning alone does not explicitly guarantee that the generated
trajectory follows the predicted structure. We therefore calibrate the base
action according to their spectral discrepancy:
\begin{equation}
    \widehat{\mathbf{A}}
    =
    \mathbf{A}_{\mathrm{base}}
    +
    \tanh(\rho)
    \operatorname{IDCT}
    \left[
        \mathbf{C}^{s}
        -
        \operatorname{DCT}(\mathbf{A}_{\mathrm{base}})
    \right].
    \label{eq:spectral_action_calibration}
\end{equation}

\rev{The spectral residual measures the discrepancy between the observation-conditioned structure prediction and the frequency representation of the base action. Through IDCT, this discrepancy is mapped back to action space to provide a lightweight correction to the generated trajectory. The learnable scalar $\rho$ controls the correction magnitude and is initialized to zero, yielding $\widehat{\mathbf{A}}=\mathbf{A}_{\mathrm{base}}$ at the beginning of training. The structure head thus provides guidance for both action generation and subsequent calibration, without introducing an additional action-generation network.}

The calibrated action is directly supervised by
\begin{equation}
    \mathcal{L}_{\mathrm{act}}
    =
    \operatorname{SmoothL1}
    (\widehat{\mathbf{A}},\mathbf{x}).
    \label{eq:action_loss}
\end{equation}
We additionally retain the dispersive feature regularizer
$\mathcal{L}_{\mathrm{dis}}$ from MP1~\citep{sheng2026mp1}, which acts on
intermediate U-Net representations. Its formulation is provided in
Appendix~\ref{app:dispersive_details}.

\subsection{MeanFlow Consistency and Endpoint Alignment}
\label{sec:velocity_space}

The calibrated action is further constrained through MeanFlow during training.
Rather than predicting an independent velocity field, SOAR derives the
corresponding transport velocity directly from the action:
\begin{equation}
    \mathbf{u}_{\theta}
    (\mathbf{z}_t,t,r,\widetilde{\mathbf{c}})
    =
    \frac{
        \mathbf{z}_t-\widehat{\mathbf{A}}
    }{
        \max(t,\varepsilon_t)
    },
    \qquad
    \varepsilon_t=10^{-5}.
    \label{eq:induced_velocity}
\end{equation}
Here, $u_\theta$ denotes the generative transport velocity rather
than the robot's physical velocity.
\rev{We adapt the data-prediction parameterization of pixel
MeanFlow~\citep{lu2026pixelmeanflow} to observation-conditioned
action generation.}

\noindent\textbf{MeanFlow consistency.}
We follow the standard random-interval MeanFlow training objective
$\mathcal{L}_{\mathrm{MF}}$
~\citep{geng2025meanflow,sheng2026mp1} to constrain the induced transport
field. The interval sampling, self-guided target, JVP computation, and
AdaptiveL2 regression follow the standard formulation and are summarized in
Appendix~\ref{app:meanflow_details}.

\noindent\textbf{Inference-endpoint alignment.}
MeanFlow is trained over random $(t,r)$ intervals, whereas one-step inference
always operates at the fixed boundary $(t,r)=(1,0)$. We therefore evaluate the
same generator under this condition and explicitly align its prediction with
the demonstrated action:
\begin{equation}
    \mathcal{L}_{\mathrm{end}}
    =
    \operatorname{AdaptiveL2}
    \left(
        \mathbf{x}
        -
        \widehat{\mathbf{A}}_{\mathrm{end}}
    \right),
    \label{eq:endpoint_loss}
\end{equation}
where $\widehat{\mathbf{A}}_{\mathrm{end}}$ is the calibrated action predicted
under the actual one-step inference condition.
This objective is equivalent to matching the corresponding endpoint transport
velocities at $t=1$; the derivation is given in
Appendix~\ref{app:endpoint_details}.

At inference, SOAR predicts $\mathbf{C}^{s}$ from the observation, constructs
$\widetilde{\mathbf{c}}$, samples a Gaussian action latent, and evaluates
$F_{\theta}(\boldsymbol{\epsilon},1,0,\widetilde{\mathbf{c}})$ once.
The resulting base action is then calibrated using ~\eqref{eq:spectral_action_calibration}.
No JVP, iterative denoising, or numerical ODE integration is required,
yielding $\mathrm{NFE}=1$.

\section{Experiments}

\subsection{Simulation Experiment With 3D Inputs}

\noindent\textbf{Experimental Setup.}
We evaluate SOAR on two widely used 3D robotic manipulation benchmarks,
Adroit~\citep{rajeswaran2018dexterous} and
Meta-World~\citep{yu2020metaworld}, following the evaluation protocols of
prior work~\citep{fang2026omp,su2025freqpolicy}. We consider two settings:
37 tasks consisting of 3 Adroit and 34 Meta-World tasks, and 53 tasks
consisting of 3 Adroit and all 50 Meta-World tasks. For each task, we use
10 expert demonstrations for training and downsample the point cloud to
512 points. Each Adroit task is trained independently for 3,000 epochs, while each
Meta-World task is trained for 1,000 epochs, using three random seeds
(0, 1, and 2). All training and evaluation are conducted on
a single NVIDIA RTX 4090 GPU. We use a batch size of 128 and optimize the
policy with AdamW using a learning rate of $1\times10^{-4}$. The
observation window and prediction horizon are set to 2 and 4,
respectively. We evaluate the policy every 200 epochs. For each random seed, the five
highest evaluation success rates are selected and averaged, and the final
task performance is reported as the mean and sample standard deviation
across the three seeds. SOAR performs one-step action generation at
inference time with $\mathrm{NFE}=1$.

\paragraph{Results.}
Tables~\ref{tab:37_benchmark_results} and
\ref{tab:53_benchmark_results} summarize the simulation results.
SOAR achieves 84.4\% and 83.9\% overall success on
the 37-task and 53-task benchmarks, respectively, while retaining
$\mathrm{NFE}=1$. The small performance change after expanding the
evaluation from 37 to 53 tasks indicates that SOAR remains robust under
broader task coverage. On the full 53-task benchmark, SOAR reaches 84.1\% on Meta-World and
shows particularly strong performance on the Medium and Hard subsets,
outperforming the best baselines by 8.5\% and 7.4\%,
respectively. This suggests that the advantage of SOAR is not limited to
easier tasks near saturation, but extends to more challenging manipulation
settings. 

The latency results in Table~\ref{tab:53_benchmark_results} and
Fig.~\ref{fig:teaser} show that these gains are obtained while retaining
one-step inference. SOAR requires only 8.7\,ms of inference latency,
remaining close to the fastest one-step baseline, MP1, while improving
overall success by 8.0\%. Compared with DP3, SOAR reduces
inference latency from 70.7\,ms to 8.7\,ms while increasing overall
success from 72.3\% to 83.9\%. These results show that SOAR improves
one-step action generation without relying on iterative refinement,
providing a favorable performance--efficiency trade-off.
Detailed per-task and reproduced results are reported in
Appendix~\ref{sec:3D}.

\begin{table}[tbp]
    \centering
    \caption{
        Performance on the 37-task benchmark with 3D inputs.
        We evaluate 37 challenging tasks over three random seeds and report
        the mean success rate (\%) and standard deviation.
        Results marked with $*$ are reproduced.
        \textbf{Bold} denotes the best result.
    }
    \vspace{3mm}
    \label{tab:37_benchmark_results}

    \setlength{\tabcolsep}{4.5pt}
    \renewcommand{\arraystretch}{1.10}

    \resizebox{\textwidth}{!}{%
    \begin{tabular}{lccccccccccc}
        \toprule

        \textbf{}
        & \textbf{}
        & \multicolumn{4}{c}{\textbf{Adroit (3)}}
        & \multicolumn{5}{c}{\textbf{Meta-World (34)}}
        & \textbf{} \\

        \cmidrule(lr){3-6}
        \cmidrule(lr){7-11}

        \textbf{Methods}
        & \textbf{NFE}
        & \textbf{Hammer}
        & \textbf{Door}
        & \textbf{Pen}
        & \textbf{Average}
        & \textbf{Easy (21)}
        & \textbf{Medium (4)}
        & \textbf{Hard (4)}
        & \textbf{Very Hard (5)}
        & \textbf{Average}
        & \textbf{Overall} \\

        \midrule

        DP~\citep{chi2023diffusion}
        & 10
        & $16.0 \pm 10.0$
        & $34.0 \pm 11.0$
        & $13.0 \pm 2.0$
        & $21.0$
        & $50.7 \pm 6.1$
        & $11.0 \pm 2.5$
        & $5.25 \pm 2.5$
        & $22.0 \pm 5.0$
        & $36.5$
        & $35.2 \pm 5.3$ \\

        DP3~\citep{ze2024dp3}
        & 10
        & $100.0 \pm 0.0$
        & $56.0 \pm 5.0$
        & $46.0 \pm 10.0$
        & $67.3$
        & $87.3 \pm 2.2$
        & $44.5 \pm 8.7$
        & $32.7 \pm 7.7$
        & $39.4 \pm 9.0$
        & $68.8$
        & $68.7 \pm 4.7$ \\

        Simple DP3~\citep{ze2024dp3}
        & 10
        & $98.0 \pm 2.0$
        & $40.0 \pm 17.0$
        & $36.0 \pm 4.0$
        & $58.0$
        & $86.8 \pm 2.3$
        & $42.0 \pm 6.5$
        & $38.7 \pm 7.5$
        & $35.0 \pm 11.6$
        & $68.3$
        & $67.4 \pm 5.0$ \\

        AdaFlow~\citep{hu2024adaflow}
        & --
        & $45.0 \pm 11.0$
        & $27.0 \pm 6.0$
        & $18.0 \pm 6.0$
        & $30.0$
        & $49.4 \pm 6.8$
        & $12.0 \pm 5.0$
        & $5.75 \pm 4.4$
        & $24.0 \pm 4.8$
        & $36.1$
        & $35.6 \pm 6.1$ \\

        CP~\citep{prasad2024consistencypolicy}
        & 1
        & $45.0 \pm 4.0$
        & $31.0 \pm 10.0$
        & $13.0 \pm 6.0$
        & $29.7$
        & $69.3 \pm 4.2$
        & $21.2 \pm 6.0$
        & $17.5 \pm 3.9$
        & $30.0 \pm 4.9$
        & $51.8$
        & $50.1 \pm 4.7$ \\

        FlowPolicy~\citep{zhang2025flowpolicy}
        & 1
        & $98.0 \pm 1.0$
        & $61.0 \pm 2.0$
        & $54.0 \pm 4.0$
        & $71.0$
        & $84.8 \pm 2.2$
        & $58.2 \pm 7.9$
        & $40.2 \pm 4.5$
        & $52.2 \pm 5.0$
        & $71.6$
        & $71.6 \pm 3.5$ \\

        MP1~\citep{sheng2026mp1}
        & 1
        & $100.0 \pm 0.0$
        & $69.0 \pm 2.0$
        & $58.0 \pm 5.0$
        & $75.7$
        & $88.2 \pm 1.1$
        & $68.0 \pm 3.1$
        & $58.1 \pm 5.0$
        & $67.2 \pm 2.7$
        & $79.2$
        & $78.9 \pm 2.1$ \\

        OMP~\citep{fang2026omp}
        & 1
        & $100.0 \pm 0.0$
        & $68.0 \pm 3.0$
        & $60.0 \pm 4.0$
        & $76.0$
        & $89.7 \pm 0.7$
        & $\mathbf{77.4 \pm 2.2}$
        & $62.5 \pm 3.1$
        & $\mathbf{77.8 \pm 3.0}$
        & $83.3$
        & $82.3 \pm 1.6$ \\

        \midrule

        \textbf{SOAR (Ours)}
        & \textbf{1}
        & $\mathbf{100.0 \pm 0.0}$
        & $\mathbf{77.5 \pm 2.4}$
        & $\mathbf{61.9 \pm 7.3}$
        & $\mathbf{79.8 \pm 2.9}$
        & $\mathbf{92.3 \pm 0.5}$
        & $76.8 \pm 0.7$
        & $\mathbf{63.5 \pm 2.2}$
        & $76.9 \pm 1.8$
        & $\mathbf{84.8 \pm 0.7}$
        & $\mathbf{84.4 \pm 0.5}$ \\

        \bottomrule
    \end{tabular}%
    }
\end{table}

\begin{table}[tbp]
    \centering
    \caption{
        Performance on the 53-task benchmark with 3D inputs.
        We evaluate 53 challenging tasks over three random seeds and report
        the mean success rate (\%) and standard deviation.
        Average inference latency is reported in milliseconds.
        Results marked with $*$ are reproduced.
        \textbf{Bold} denotes the best result.
    }
    \label{tab:53_benchmark_results}
    \vspace{3mm}

    \setlength{\tabcolsep}{3.8pt}
    \renewcommand{\arraystretch}{1.10}

    \resizebox{\textwidth}{!}{%
    \begin{tabular}{lcccccccccccc}
        \toprule

        \textbf{}
        & \textbf{}
        & \multicolumn{4}{c}{\textbf{Adroit (3)}}
        & \multicolumn{5}{c}{\textbf{Meta-World (50)}}
        & \textbf{}
        & \textbf{} \\

        \cmidrule(lr){3-6}
        \cmidrule(lr){7-11}

        \textbf{Methods}
        & \textbf{NFE}
        & \textbf{Hammer}
        & \textbf{Door}
        & \textbf{Pen}
        & \textbf{Average}
        & \textbf{Easy (28)}
        & \textbf{Medium (11)}
        & \textbf{Hard (6)}
        & \textbf{Very Hard (5)}
        & \textbf{Average}
        & \textbf{Overall}
        & \textbf{Inference Latency (ms)} \\

        \midrule

        DP$^{*}$~\citep{chi2023diffusion}
        & 100
        & $75.1 \pm 3.5$
        & $23.7 \pm 3.7$
        & $19.3 \pm 1.8$
        & $39.4 \pm 1.5$
        & $69.0 \pm 1.6$
        & $32.8 \pm 2.4$
        & $34.3 \pm 2.9$
        & $31.5 \pm 5.7$
        & $53.1 \pm 1.1$
        & $52.3 \pm 0.9$
        & $678.4 \pm 6.9$ \\

        DP3$^{*}$~\citep{ze2024dp3}
        & 10
        & $100.0 \pm 0.0$
        & $73.5 \pm 2.1$
        & $56.7 \pm 2.6$
        & $76.7 \pm 1.6$
        & $83.1 \pm 0.7$
        & $58.7 \pm 3.6$
        & $54.1 \pm 3.9$
        & $60.7 \pm 1.5$
        & $72.0 \pm 1.5$
        & $72.3 \pm 1.4$
        & $70.7 \pm 2.6$ \\

        Simple DP3$^{*}$~\citep{ze2024dp3}
        & 10
        & $100.0 \pm 0.0$
        & $56.8 \pm 1.2$
        & $56.1 \pm 0.3$
        & $71.0 \pm 0.4$
        & $82.0 \pm 1.1$
        & $53.8 \pm 3.1$
        & $54.3 \pm 5.3$
        & $61.7 \pm 4.5$
        & $70.4 \pm 1.6$
        & $70.4 \pm 1.5$
        & $43.4 \pm 0.4$ \\

        ManiCM~\citep{gao2026manicm}
        & 1
        & $100.0 \pm 0.0$
        & $68.0 \pm 1.0$
        & $49.0 \pm 4.0$
        & $72.3$
        & $83.6$
        & $55.6$
        & $33.3$
        & $67.0$
        & $69.7$
        & $69.0 \pm 4.6$
        & -- \\

        SDM~\citep{jia2024sdm}
        & 1
        & $100.0 \pm 0.0$
        & $73.0 \pm 2.0$
        & $49.0 \pm 4.0$
        & $74.0$
        & $86.5$
        & $65.8$
        & $35.8$
        & $71.6$
        & $74.4$
        & $74.8 \pm 4.5$
        & -- \\

        FlowPolicy$^{*}$~\citep{zhang2025flowpolicy}
        & 1
        & $95.2 \pm 2.4$
        & $57.3 \pm 6.2$
        & $47.8 \pm 11.0$
        & $66.8 \pm 6.4$
        & $90.6 \pm 0.5$
        & $56.0 \pm 1.0$
        & $54.4 \pm 11.3$
        & $62.7 \pm 2.4$
        & $75.9 \pm 1.6$
        & $75.3 \pm 1.8$
        & $14.7 \pm 0.0$ \\

        FreqPolicy~\citep{su2025freqpolicy}
        & 1
        & $98.0 \pm 2.0$
        & $68.0 \pm 5.0$
        & $52.0 \pm 4.0$
        & $72.7$
        & $90.6$
        & $65.8$
        & $46.8$
        & $\mathbf{79.4}$
        & $78.8$
        & $78.5 \pm 2.6$
        & -- \\

        MP1$^{*}$~\citep{sheng2026mp1}
        & 1
        & $96.1 \pm 2.3$
        & $69.2 \pm 3.1$
        & $57.1 \pm 6.1$
        & $74.1 \pm 2.0$
        & $88.5 \pm 0.8$
        & $59.5 \pm 2.9$
        & $56.1 \pm 3.2$
        & $65.7 \pm 1.3$
        & $76.0 \pm 1.5$
        & $75.9 \pm 1.3$
        & $\mathbf{8.1 \pm 0.0}$ \\

        \midrule

        \textbf{SOAR (Ours)}
        & \textbf{1}
        & $\mathbf{100.0 \pm 0.0}$
        & $\mathbf{77.5 \pm 2.4}$
        & $\mathbf{61.9 \pm 7.3}$
        & $\mathbf{79.8 \pm 2.9}$
        & $\mathbf{93.7 \pm 0.7}$
        & $\mathbf{74.3 \pm 0.6}$
        & $\mathbf{63.5 \pm 1.8}$
        & $76.9 \pm 1.8$
        & $\mathbf{84.1 \pm 0.5}$
        & $\mathbf{83.9 \pm 0.3}$
        & $8.7 \pm 0.1$ \\

        \bottomrule
    \end{tabular}%
    }
\end{table}


\rev{

\subsection{Simulation Experiments with VLA Settings}
\label{sec:vla_simulation}

\paragraph{Experimental Setup.}
We further investigate the transferability of SOAR
from 3D visuomotor policies to large-scale
vision-language-action (VLA) models.
Specifically, we replace the original action expert
of $\pi_{0.5}$~\citep{black2025pi05} with SOAR
and evaluate the resulting policy on
LIBERO~\citep{liu2023libero} and
RoboTwin 2.0~\citep{chen2025robotwin}. For both benchmarks, the vision-language backbone
is initialized from the OpenPI
\texttt{pi05\_base} checkpoint.
On LIBERO, we jointly train a single policy
over all 40 tasks from the four standard suites.
SOAR predicts 10-step action chunks and executes
five actions before replanning. On RoboTwin 2.0, we jointly train a single policy
over all 50 tasks using clean demonstrations.
The $\pi_{0.5}$-SOAR is trained for 100K optimization steps. SOAR predicts 50-step joint-position action chunks. In both settings, SOAR generates the action chunk
with a single action-expert evaluation
($\mathrm{NFE}=1$).
For direct comparison, we additionally reproduce
$\pi_{0.5}$ baselines with different numbers of
action-generation evaluations.
These reproduced results are marked with
a superscript asterisk ($^{*}$) in the tables,
while unmarked baseline results are taken from
the corresponding references.
Detailed training and evaluation configurations
are provided in Appendix~\ref{sec:vla_details}.

\noindent\textbf{Results.}
Table~\ref{tab:libero_comparison} presents the
results on LIBERO.
With $\mathrm{NFE}=1$, $\pi_{0.5}$-SOAR achieves
an average success rate of 97.2\%,
obtaining 99.8\% on Spatial, 97.4\% on Object,
98.5\% on Goal, and 93.1\% on Long. Our reproduced $\pi_{0.5}$ baselines achieve
average success rates of 96.5\%, 96.1\%,
and 95.5\% with $\mathrm{NFE}=10$, $2$,
and $1$, respectively.
SOAR improves the average success rate by
0.7 percentage points over the reproduced
10-step baseline and by 1.7 percentage points
over its one-step counterpart, while requiring
only one action-expert evaluation.
It also achieves the highest average success
rate among the compared methods with
$\mathrm{NFE}\leq 2$ and remains competitive
with higher-NFE VLA policies.

We further evaluate SOAR on RoboTwin 2.0,
which involves 50 jointly trained manipulation
tasks and substantially longer action chunks.
As shown in Table~\ref{tab:robotwin_comparison},
$\pi_{0.5}$-SOAR achieves an average success
rate of 71.7\% under the clean-to-clean setting
with $\mathrm{NFE}=1$.
Our reproduced $\pi_{0.5}$ baselines achieve
70.0\%, 69.1\%, and 65.1\% with
$\mathrm{NFE}=10$, $2$, and $1$, respectively.
SOAR therefore exceeds the reproduced one-step
baseline by 6.6 percentage points and the
10-step baseline by 1.7 percentage points. Across LIBERO and RoboTwin 2.0, SOAR
consistently outperforms the reproduced one-step
$\pi_{0.5}$ baselines while maintaining
competitive performance against multi-step
generation.
These results demonstrate the effectiveness
and transferability of SOAR, showing that
its structure-guided one-step generation
can be successfully integrated into VLA
policies across different manipulation
benchmarks while preserving high task
performance.

\begin{table}[tbp]
    \centering
    \caption{
        Comparison with representative VLA methods on LIBERO.
        Results are success rates (\%).
        \textsuperscript{*} denotes baseline results
        reproduced and evaluated in this work;
        unmarked baseline results are taken from
        the corresponding references.
        NFE denotes the number of action-expert
        evaluations per policy inference.
        \textbf{Bold} denotes the best result
        within each comparison block.
    }
    \label{tab:libero_comparison}
    \vspace{2mm}

    \setlength{\tabcolsep}{14pt}
    \renewcommand{\arraystretch}{1.10}

    \resizebox{\textwidth}{!}{%
    \begin{tabular}{lccccccc}
        \toprule
        \textbf{Methods}
        & \textbf{Publication}
        & \textbf{NFE}
        & \textbf{Spatial}
        & \textbf{Object}
        & \textbf{Goal}
        & \textbf{Long}
        & \textbf{Average} \\
        \midrule

        UniVLA~\citep{bu2025univla}
        & RSS, 2025 & AR
        & 96.5 & 96.8 & 95.6 & 92.0 & 95.2 \\

        OpenVLA-OFT~\citep{kim2025openvlaoft}
        & RSS, 2025 & AR
        & 97.6 & 98.4 & 97.9 & 94.5 & 97.1 \\

        VLA-Cache~\citep{xu2025vlacache}
        & NeurIPS, 2025 & AR
        & \textbf{98.3} & 97.5
        & \textbf{98.3} & \textbf{95.4}
        & \textbf{97.4} \\

        UnifiedVLA~\citep{wang2026unifiedvla}
        & ICLR, 2026 & AR
        & 95.4 & \textbf{98.8}
        & 93.6 & 94.0 & 95.5 \\

        \midrule

        Dita~\citep{hou2025dita}
        & ICCV, 2025 & 20
        & 84.2 & 96.3 & 85.4 & 63.8 & 82.4 \\

        CronusVLA~\citep{li2026cronusvla}
        & AAAI, 2026 & 100
        & \textbf{97.3} & \textbf{99.6}
        & \textbf{96.9} & \textbf{94.0}
        & \textbf{97.0} \\

        \midrule

        DreamVLA~\citep{zhang2025dreamvla}
        & NeurIPS, 2025 & 10
        & 97.5 & 94.0 & 89.5 & 89.5 & 92.6 \\

        $\pi_{0.5}$\textsuperscript{*}~
        \citep{black2025pi05}
        & CoRL, 2025 & 10
        & 98.6 & 97.9 & 98.0 & 91.5 & 96.5 \\

        FLOWER~\citep{reuss2025flower}
        & CoRL, 2025 & 8
        & 97.5 & 99.1 & 96.1 & 94.9 & 96.9 \\

        $\pi_{0.5}$~\citep{black2025pi05}
        & CoRL, 2025 & 10
        & \textbf{98.8} & 98.2
        & 98.0 & 92.4 & 96.9 \\

        InstructVLA~\citep{yang2026instructvla}
        & ICLR, 2026 & 10
        & 97.3 & 99.6 & 96.5 & 89.8 & 95.8 \\

        MemoryVLA~\citep{shi2026memoryvla}
        & ICLR, 2026 & 10
        & 98.4 & 98.4 & 96.4 & 93.4 & 96.5 \\

        X-VLA~\citep{zheng2026xvla}
        & ICLR, 2026 & 10
        & 98.2 & 98.6 & 97.8
        & \textbf{97.6} & 98.1 \\

        Cosmos Policy~\citep{kim2026cosmospolicy}
        & ICLR, 2026 & 10
        & 98.1 & \textbf{100.0}
        & \textbf{98.2} & \textbf{97.6}
        & \textbf{98.5} \\

        \midrule

        UniAct~\citep{zheng2025uniact}
        & CVPR, 2025 & 1
        & 77.0 & 87.0 & 77.0 & 70.0 & 77.8 \\

        MolmoAct-7B-D~\citep{lee2025molmoact}
        & arXiv, 2025 & 1
        & 87.0 & 95.4 & 87.6 & 77.2 & 86.6 \\

        NinA (MLP)~\citep{tarasov2025nina}
        & NeurIPS WS, 2025 & 1
        & 87.8 & 98.2 & 90.2 & 92.8 & 90.9 \\

        EveryDayVLA~\citep{chopra2025everydayvla}
        & ICRA subm., 2026 & 1
        & 96.8 & 95.6 & 91.0 & 82.0 & 91.4 \\

        $\pi_{0.5}$\textsuperscript{*}~
        \citep{black2025pi05}
        & CoRL, 2025 & 1
        & 97.9 & 96.0 & 97.7 & 90.3 & 95.5 \\

        $\pi_{0.5}$\textsuperscript{*}~
        \citep{black2025pi05}
        & CoRL, 2025 & 2
        & 98.5 & 97.6 & 97.4 & 90.9 & 96.1 \\

        PI-VLA~\citep{jian2026pivla}
        & Symmetry, 2026 & 1
        & 79.5 & 73.4 & 73.3 & 66.6 & 73.2 \\

        MIP ($\pi_{0}$ arch.)~\citep{pan2026mip}
        & ICLR, 2026 & 2
        & 97.6 & 95.8 & 95.2 & 82.2 & 92.7 \\

        CF-VLA~\citep{du2026cfvla}
        & ACM MM, 2026 & 2
        & 98.0 & \textbf{99.2}
        & 96.6 & 92.0 & 96.5 \\

        \midrule

        \textbf{$\pi_{0.5}$-SOAR}
        & -- & \textbf{1}
        & \textbf{99.8}
        & 97.4
        & \textbf{98.5}
        & \textbf{93.1}
        & \textbf{97.2} \\

        \bottomrule
    \end{tabular}%
    }
\end{table}

\begin{table}[tbp]
    \centering
    \caption{
        Comparison with representative methods
        on RoboTwin 2.0 under the clean-to-clean setting.
        Results are average success rates (\%)
        across 50 tasks.
        \textsuperscript{*} denotes baseline results
        reproduced and evaluated in this work;
        unmarked baseline results are taken from
        the corresponding references.
        NFE denotes the number of action-expert
        evaluations per policy inference.
        \textbf{Bold} denotes the best result
        within each comparison block.
    }
    \label{tab:robotwin_comparison}
    \vspace{2mm}

    \setlength{\tabcolsep}{21pt}
    \renewcommand{\arraystretch}{1.10}

    \resizebox{0.8\columnwidth}{!}{%
    \begin{tabular}{lcccc}
        \toprule
        \textbf{Methods}
        & \textbf{Publication}
        & \textbf{Params}
        & \textbf{NFE}
        & \textbf{Success} \\
        \midrule

        Diffusion Policy~\citep{chi2023diffusion}
        & RSS, 2023 & 0.1B & 100 & 28.0 \\

        SeedPolicy~\citep{gui2026seedpolicy}
        & arXiv, 2026 & 0.2B & 100
        & \textbf{42.8} \\

        \midrule

        DP3~\citep{ze2024dp3}
        & RSS, 2024 & 0.3B & 10 & 55.2 \\

        $\pi_{0}$~\citep{black2025pi0}
        & RSS, 2025 & 3.2B & 10 & 46.4 \\

        $\pi_{0.5}$\textsuperscript{*}~
        \citep{black2025pi05}
        & CoRL, 2025 & 3.4B & 10 & 70.0 \\

        $\pi_{0.5}$~\citep{black2025pi05}
        & CoRL, 2025 & 3.4B & 10
        & \textbf{70.7} \\

        \midrule

        ACT~\citep{zhao2023act}
        & RSS, 2023 & 0.1B & 1 & 29.7 \\

        RDT~\citep{liu2025rdt}
        & ICLR, 2025 & 1.7B & 5 & 34.5 \\

        FlowPolicy~\citep{zhang2025flowpolicy}
        & AAAI, 2025 & 0.3B & 1 & 41.0 \\

        UP-VLA~\citep{zhang2025upvla}
        & ICML, 2025 & 1.6B & 1 & 52.9 \\

        $\pi_{0.5}$\textsuperscript{*}~
        \citep{black2025pi05}
        & CoRL, 2025 & 3.4B & 1 & 65.1 \\

        $\pi_{0.5}$\textsuperscript{*}~
        \citep{black2025pi05}
        & CoRL, 2025 & 3.4B & 2 & 69.1 \\

        StarVLA-$\alpha$~\citep{ye2026starvla}
        & ECCV, 2026 & 3.8B & 1 & 50.3 \\

        \midrule

        \textbf{$\pi_{0.5}$-SOAR}
        & -- & 2.9B & \textbf{1}
        & \textbf{71.7} \\

        \bottomrule
    \end{tabular}%
    }
\end{table}

}

\rev{
\subsection{Real-World Experiments}
\label{sec:real_world}

\noindent\textbf{Experimental Setup.}
We evaluate $\pi_{0.5}$~\citep{black2025pi05} and
$\pi_{0.5}$-SOAR on two real-world manipulation tasks.
The first is a long-horizon Tidy-up task, where the robot sorts
multiple tabletop objects by discarding trash into a trash bin
and placing other objects into a storage basket.
We evaluate 40 object configurations, each containing two trash
items and two non-trash objects.
The second is a dynamic pick-and-place task, where the target
object is displaced four times during grasping, with 20 trials
per model.

All models use the same observations, action representation,
and deployment platform.
We compare $\pi_{0.5}$ with 10 and 1 inference steps against
$\pi_{0.5}$-SOAR, which generates a complete 16-step action
chunk with a single network evaluation ($\mathrm{NFE}=1$).
We report task success and inference latency for Tidy-up,
and placement success and empty grasps per trial for dynamic
pick-and-place.
Further details are provided in
Appendix~\ref{app:real_world_details}.

\begin{figure}[t]
    \centering
    \includegraphics[width=\textwidth]{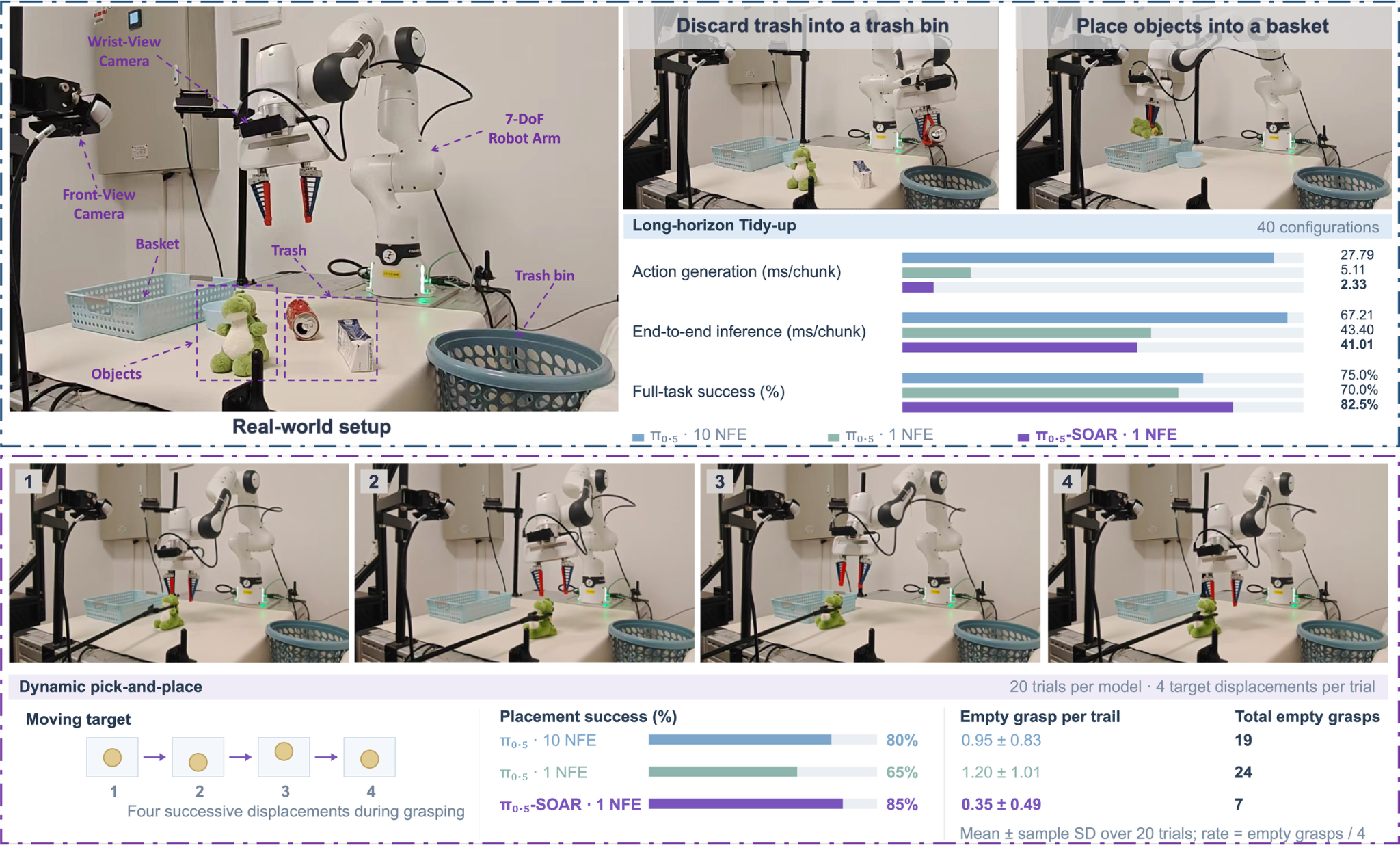}
    \caption{
    Real-world evaluation of $\pi_{0.5}$ and $\pi_{0.5}$-SOAR.
    \textbf{Top left:} Experimental setup with a 7-DoF robot arm,
    wrist-view and front-view cameras, tabletop objects,
    a storage basket, and a trash bin.
    \textbf{Top right:} Example executions of the long-horizon
    Tidy-up task, with full-task success and inference latency
    evaluated over 40 configurations.
    \textbf{Bottom:} Dynamic pick-and-place with four successive
    target displacements during grasping, evaluated over 20
    trials per model.
    Results include placement success, empty grasps per trial
    (mean $\pm$ sample standard deviation), and total empty grasps.
    }
    \label{fig:real_world_results}
\end{figure}

\noindent\textbf{Results.}
Figure~\ref{fig:real_world_results} presents the real-world results.
On the long-horizon Tidy-up task, $\pi_{0.5}$-SOAR achieves
82.5\% success, exceeding the 10-step and 1-step $\pi_{0.5}$
baselines by 7.5 and 12.5 percentage points, respectively.
Compared with the 10-step baseline, SOAR reduces action-generation
latency from 27.79\,ms to 2.33\,ms and end-to-end
inference latency from 67.21\,ms to 41.01\,ms.
Under the same one-step setting, SOAR further reduces
action-generation latency by 54.4\% while improving task
success from 70.0\% to 82.5\%.
Although directly reducing the original $\pi_{0.5}$ from
10 to 1 inference step lowers its success rate, SOAR achieves
higher success than both baselines with lower inference latency.

On dynamic pick-and-place, $\pi_{0.5}$-SOAR achieves 85\%
placement success, compared with 80\% and 65\% for the
10-step and 1-step baselines, respectively.
It also records fewer empty grasps per trial
($0.35\pm0.49$), compared with $0.95\pm0.83$ and
$1.20\pm1.01$ for the two baselines.
The corresponding total empty-grasp counts are 7, 19, and 24.
Relative to the 10-step baseline, SOAR reduces the average
number of empty grasps by 63.2\% while achieving higher
placement success.
These results indicate that SOAR retains reliable manipulation
under repeated target displacements despite using only one
network evaluation for action generation.
}

\rev{\subsection{Ablation Studies}
\label{sec:ablation}

\noindent\textbf{Ablation Setup.}
We evaluate SOAR on 20 manipulation tasks, including
three Adroit tasks (Hammer, Door, and Pen) and
17 Meta-World tasks (11 Medium and 6 Hard).
Each task is trained independently using 10 expert
demonstrations and three random seeds, with 3,000
epochs for Adroit and 1,000 epochs for Meta-World.
Following the main evaluation protocol, we average the
five highest periodic evaluation success rates for each
task and seed, and report the mean and sample standard
deviation across seeds.

We examine four components: structure supervision (SS),
multi-band conditioning (MC), action calibration (AC),
and inference-endpoint alignment (EA). 
SS trains an auxiliary head to predict key structural
information of future action trajectories from observations,
rather than serving as an independent action predictor.
 MC incorporates the predicted frequency
components into the generation condition, while AC
uses the same structure estimate to correct the
generated actions. EA aligns training with the
one-step inference condition. We progressively add
these components to a direct-action baseline and
further evaluate variants without MC or AC.
We also replace spectral calibration with direct
action-space correction and compare direct clean-action
prediction ($x$-prediction) with direct MeanFlow
velocity prediction ($u$-prediction), where the base
action is recovered from the predicted velocity
before calibration. All variants retain one-step
inference ($\mathrm{NFE}=1$).

\begin{table}[tbp]
    \centering
    \caption{
        Component ablations on Adroit and Meta-World.
        Success rates (\%) are reported as mean $\pm$
        sample standard deviation over three training seeds,
        using the five highest periodic evaluations per
        task and seed.
        SS: structure supervision; MC: multi-band conditioning;
        AC: action calibration; EA: inference-endpoint alignment.
        In the AC column, $\checkmark$ denotes spectral
        calibration, while Direct denotes action-space correction.
        The $u$-prediction variant directly predicts MeanFlow
        transport velocity instead of clean actions, recovering
        the base action before calibration.
        Average is the task-weighted mean over 20 tasks,
        computed from the displayed group means.
        ``--'' denotes an absent component.
        Bold indicates the highest mean in the Door, Pen,
        Medium, Hard, and Average columns.
    }
    \label{tab:ablation_adroit}
    \vspace{2mm}

    \setlength{\tabcolsep}{5pt}
    \renewcommand{\arraystretch}{1.12}

    \resizebox{\textwidth}{!}{%
    \begin{tabular}{lcccccccccc}
        \toprule
        & & & & &
        \multicolumn{3}{c}{\textbf{Adroit (3)}}
        & \multicolumn{2}{c}{\textbf{Meta-World (17)}}
        & \\
        \cmidrule(lr){6-8}
        \cmidrule(lr){9-10}

        \textbf{Variant}
        & \textbf{SS}
        & \textbf{MC}
        & \textbf{AC}
        & \textbf{EA}
        & \textbf{Hammer}
        & \textbf{Door}
        & \textbf{Pen}
        & \textbf{Medium (11)}
        & \textbf{Hard (6)}
        & \textbf{Average} \\
        \midrule

        Direct-action baseline
        & -- & -- & -- & --
        & $100.0 \pm 0.0$
        & $73.6 \pm 4.1$
        & $58.2 \pm 4.3$
        & $71.7 \pm 1.3$
        & $60.9 \pm 2.6$
        & $69.3$ \\

        + SS
        & $\checkmark$ & -- & -- & --
        & $100.0 \pm 0.0$
        & $74.1 \pm 4.3$
        & $61.1 \pm 5.5$
        & $71.7 \pm 0.2$
        & $60.9 \pm 1.9$
        & $69.5$ \\

        + SS + MC
        & $\checkmark$ & $\checkmark$ & -- & --
        & $100.0 \pm 0.0$
        & $75.4 \pm 5.1$
        & $59.7 \pm 5.3$
        & $73.0 \pm 0.4$
        & $61.9 \pm 2.9$
        & $70.5$ \\

        + SS + MC + AC
        & $\checkmark$ & $\checkmark$
        & $\checkmark$ & --
        & $100.0 \pm 0.0$
        & $77.1 \pm 2.0$
        & $59.3 \pm 6.2$
        & $73.6 \pm 0.5$
        & $62.4 \pm 2.3$
        & $71.0$ \\

        \midrule

        \textbf{SOAR}
        & $\checkmark$ & $\checkmark$
        & $\checkmark$ & $\checkmark$
        & $100.0 \pm 0.0$
        & $77.5 \pm 2.4$
        & $61.9 \pm 7.3$
        & $\mathbf{74.3 \pm 0.6}$
        & $63.5 \pm 1.8$
        & $\mathbf{71.9}$ \\

        \midrule

        SOAR w/o MC
        & $\checkmark$ & --
        & $\checkmark$ & $\checkmark$
        & $100.0 \pm 0.0$
        & $76.7 \pm 5.0$
        & $58.7 \pm 4.9$
        & $72.9 \pm 1.3$
        & $62.9 \pm 1.9$
        & $70.7$ \\

        SOAR w/o AC
        & $\checkmark$ & $\checkmark$
        & -- & $\checkmark$
        & $100.0 \pm 0.0$
        & $\mathbf{77.7 \pm 3.3}$
        & $59.5 \pm 6.3$
        & $72.3 \pm 0.9$
        & $62.2 \pm 3.0$
        & $70.3$ \\

        SOAR w/ direct correction
        & $\checkmark$ & $\checkmark$
        & Direct & $\checkmark$
        & $100.0 \pm 0.0$
        & $75.6 \pm 4.4$
        & $62.7 \pm 9.1$
        & $72.3 \pm 0.3$
        & $60.9 \pm 3.3$
        & $70.0$ \\

        SOAR w/ $u$-prediction
        & $\checkmark$ & $\checkmark$
        & $\checkmark$ & $\checkmark$
        & $99.2 \pm 0.7$
        & $75.3 \pm 2.9$
        & $\mathbf{68.3 \pm 5.6}$
        & $70.7 \pm 2.0$
        & $\mathbf{65.7 \pm 2.1}$
        & $70.7$ \\

        \bottomrule
    \end{tabular}%
    }
\end{table}

\noindent\textbf{Results.}
Table~\ref{tab:ablation_adroit} reports the results
across 20 tasks. The direct-action baseline achieves
an average success rate of 69.3\%. Adding SS alone
slightly improves the average to 69.5\%, with no
improvement on Meta-World Medium or Hard, consistent
with its role in learning key temporal structural
information rather than directly predicting executable
actions. Incorporating this information through MC
raises the average to 70.5\%, with Medium and Hard
improving from 71.7\% to 73.0\% and from 60.9\% to
61.9\%, respectively. Adding AC further increases
the average to 71.0\%, indicating that MC and AC
effectively utilize the structural information learned
through SS. Finally, EA raises the average to 71.9\%,
with Medium and Hard reaching 74.3\% and 63.5\%.
This further gain supports the benefit of aligning
training with the one-step inference condition.

Removing MC or AC from the complete model reduces
the average success rate to 70.7\% and 70.3\%,
respectively, while replacing spectral calibration
with direct action-space correction yields 70.0\%.
These results further support the contributions of
structural conditioning and spectral calibration
to action generation. We also compare action parameterizations.
Although $u$-prediction performs better on Pen
(68.3\% vs. 61.9\%) and Meta-World Hard
(65.7\% vs. 63.5\%), it underperforms on Door
(75.3\% vs. 77.5\%) and Meta-World Medium
(70.7\% vs. 74.3\%). Overall, direct clean-action
prediction achieves a higher average success rate
(71.9\% vs. 70.7\%), although the relative
advantage varies across tasks.

}

\section{Conclusion}

We presented SOAR, a structure-guided one-step policy that predicts key temporal structural information from observations to guide and calibrate clean-action generation. By correcting structural discrepancies in the generated actions, SOAR improves action quality without additional evaluations of the generative backbone. SOAR achieves 83.9\% overall success across 53 Adroit and Meta-World tasks with 8.7 ms inference latency. Its improvements over reproduced one-step baselines on LIBERO and RoboTwin 2.0, together with the gains in real-world experiments, demonstrate the effectiveness of the proposed approach and its transferability to VLA policies.

\section*{AI Use Statement}
Generative AI tools assisted with experimental interpretation, methodological feedback, manuscript drafting and editing, and literature review. The authors take full responsibility for the final content and all AI-assisted materials.

\section*{Reproducibility Statement}
Section~\ref{sec:method} specifies the SOAR architecture and training
objectives. Appendix~\ref{app:method_details} provides implementation
details for frequency-domain structure prediction and calibration,
MeanFlow training, inference-endpoint alignment, and dispersive feature
regularization. Appendix~\ref{app:learned_parameters} reports the DCT
configuration, loss weights, and learned calibration and conditioning
parameters. Section~4 describes the experimental settings and evaluation protocols,
including the training seeds and success-rate aggregation procedure
for the 3D simulations. Appendix~\ref{sec:3D} reports per-task results
on Adroit and Meta-World and specifies the inference-latency measurement
protocol. Appendix~\ref{sec:vla_details} describes model initialization,
training configurations, and evaluation protocols for LIBERO and
RoboTwin~2.0. Appendix~\ref{app:real_world_details} describes the
real-robot data, training and deployment settings, and evaluation metrics.
Tables~\ref{tab:tidy_per_configuration} and~\ref{tab:dynamic_per_trial}
report the outcomes for each Tidy-up configuration and each dynamic
pick-and-place trial, respectively.

\bibliographystyle{plainnat}
\bibliography{references}

\clearpage

\appendix
\section{Appendix}

\subsection{Additional Method Details}
\label{app:method_details}

This appendix provides additional implementation details for the
frequency-domain structure module, MeanFlow training, inference-endpoint
alignment, and dispersive feature regularization. These details are omitted
from the main text for clarity.


\subsubsection{Frequency-Domain Structure Details}

\label{app:structure_details}

\rev{The structure head consists of three linear layers with two 512-dimensional hidden layers. Mish activations follow the first two layers, and the output layer is linear. The head maps the observation condition to the predicted DCT coefficients, which are reshaped into frequency-by-action-channel format.The head receives observation-derived conditions; demonstrated future actions are used only as training targets.We apply an orthonormal DCT-II along the temporal dimension independently for each included action channel. The inverse uses the transpose of the orthonormal transform matrix. We retain the complete spectrum, with the number of coefficients equal to the prediction horizon: 4 for Adroit/Meta-World, 10 for LIBERO, 50 for RoboTwin, and 16 for the real robot. For real-robot adaptation, structure supervision, spectral conditioning, and spectral calibration operate only on the seven continuous joint channels, excluding the binary gripper.}

\noindent\textbf{Multi-band structure conditioning.}
Let $K$ denote the number of DCT coefficients represented by the structure
branch. We define the normalized frequency index as
$\nu_k = \frac{k}{K-1}$,
and partition the predicted spectrum into low-, mid-, and high-frequency
groups:
$\mathcal{B}_{L}=\{k\mid\nu_k<\frac{1}{2}\}$,
$\mathcal{B}_{M}=\{k\mid\frac{1}{2}\leq\nu_k<\frac{3}{4}\}$,
and
$\mathcal{B}_{H}=\{k\mid\nu_k\geq\frac{3}{4}\}$.
Lower-frequency coefficients primarily describe slowly varying trajectory
structure, while higher-frequency coefficients capture progressively
shorter-scale temporal variations.

The three groups are independently projected into the observation feature
space and fused with the global condition:
\begin{equation}
\begin{aligned}
    \widetilde{\mathbf{c}}
    =\mathbf{c}
    &+g_L P_L\left(\mathbf{C}^{s}_{\mathcal{B}_L}\right)
    +g_M P_M\left(\mathbf{C}^{s}_{\mathcal{B}_M}\right)
    +g_H P_H\left(\mathbf{C}^{s}_{\mathcal{B}_H}\right),
\end{aligned}
\label{eq:app_multiband_injection}
\end{equation}
where $P_L$, $P_M$, and $P_H$ are learnable projections, and
$g_L$, $g_M$, and $g_H$ are learnable scalar gates.
All gates are initialized to zero, yielding
$\widetilde{\mathbf{c}}=\mathbf{c}$ at initialization.
This initialization preserves the original observation condition at the
beginning of training and introduces structural guidance progressively.

\noindent\textbf{Adaptive structure supervision.}
DCT coefficients may differ substantially in scale. A uniform spectral
objective can therefore be dominated by high-energy components even when
lower-energy frequencies remain difficult to predict. We instead measure the
prediction difficulty of each coefficient relative to its own target energy.

\begingroup\emergencystretch=2em
For sample $b$ and frequency coefficient $k$, we define the prediction error
and target energy as
$e_{b,k}=\frac{1}{D_a}\sum_{d=1}^{D_a}
(C^{s}_{b,k,d}-C^{\mathrm{gt}}_{b,k,d})^2$,
and
$E_{b,k}=\frac{1}{D_a}\sum_{d=1}^{D_a}
(C^{\mathrm{gt}}_{b,k,d})^2$.
The relative prediction difficulty is
$d_{b,k}=\frac{e_{b,k}}{\max(E_{b,k},10^{-6})}$.
\par\endgroup

We convert the difficulty scores into normalized positive weights:
\begin{equation}
    \omega_{b,k}
    =
    \frac{
        \exp\left(\gamma_s\operatorname{sg}[d_{b,k}]\right)
    }{
        \sum_{j=0}^{K-1}
        \exp\left(\gamma_s\operatorname{sg}[d_{b,j}]\right)
    },
    \qquad \gamma_s=2,
    \label{eq:app_frequency_weight}
\end{equation}
where $\operatorname{sg}[\cdot]$ denotes stop-gradient.
The structure objective is
$\mathcal{L}_{\mathrm{struct}}
=\frac{1}{B}\sum_{b=1}^{B}
\sum_{k=0}^{K-1}\omega_{b,k}e_{b,k}$.

\rev{Thus, coefficients with larger prediction errors relative to their
own target energy receive stronger supervision.
This objective shares the adaptive spectral-reweighting principle
of FreqPolicy~\citep{su2025freqpolicy}, but differs in its learning
target.
FreqPolicy weights frequency-domain discrepancies between velocity
predictions at different flow times.
Here, the weights are computed from the error between the
observation-conditioned structure prediction $C^s$ and the
demonstration spectrum $C^{\mathrm{gt}}$, normalized by target
coefficient energy.
Moreover, the predicted spectrum is used during inference through
multi-band conditioning and action calibration.
The distinction therefore concerns both what the structure branch
learns and how its prediction participates in action generation,
rather than the use of DCT or adaptive weighting alone.}


\subsubsection{MeanFlow Training Details}
\label{app:meanflow_details}

\noindent\textbf{Linear probability path.}
During training, we sample Gaussian noise
$\boldsymbol{\epsilon}\sim\mathcal{N}(\mathbf{0},\mathbf{I})$
and construct the intermediate action state
$\mathbf{z}_t=(1-t)\mathbf{x}+t\boldsymbol{\epsilon}$,
with $t\in[0,1]$.
The path interpolates between the demonstrated clean action
$\mathbf{x}$ at $t=0$ and Gaussian noise at $t=1$.

\noindent\textbf{Interval sampling.}
Following MeanFlow~\citep{geng2025meanflow,sheng2026mp1}, we sample
$a_1,a_2\sim\mathcal{N}(-0.4,1)$,
$s_i=\sigma(a_i)$,
$t=\max(s_1,s_2)$, and
$r=\min(s_1,s_2)$,
where $\sigma(\cdot)$ denotes the sigmoid function.
This guarantees $0\leq r\leq t\leq1$.
For half of the samples in each mini-batch, we further set $r=t$,
combining finite-interval MeanFlow supervision with the zero-length
flow-matching case.

\noindent\textbf{Self-guided MeanFlow target.}
For the linear probability path, the instantaneous direction is
$\mathbf{v}=\boldsymbol{\epsilon}-\mathbf{x}$.
We evaluate the current transport field at $(t,t)$ without gradient
propagation and construct the self-guided direction
\begin{equation}
    \widehat{\mathbf{v}}
    =
    w\mathbf{v}
    +
    (1-w)
    \operatorname{sg}
    \left[
        \mathbf{u}_{\theta}
        \left(
            \mathbf{z}_t,
            t,
            t,
            \widetilde{\mathbf{c}}
        \right)
    \right],
    \qquad
    w=2.
    \label{eq:app_bootstrap_velocity}
\end{equation}

The directional derivative of the transport field is evaluated through a
Jacobian--vector product:
\begin{equation}
    \dot{\mathbf{u}}_{\theta}
    =
    \operatorname{JVP}_{(\mathbf{z},t,r)}
    \left[
        \mathbf{u}_{\theta};
        \left(
            \widehat{\mathbf{v}},
            \mathbf{1},
            \mathbf{0}
        \right)
    \right]_
    {(\mathbf{z}_t,t,r)}.
    \label{eq:app_meanflow_jvp}
\end{equation}
The tangent in the $r$ direction is set to zero.

The resulting MeanFlow target is
$\mathbf{u}_{\mathrm{tgt}}
=\operatorname{sg}[
\widehat{\mathbf{v}}-(t-r)\dot{\mathbf{u}}_{\theta}]$.
The corresponding objective is
$\mathcal{L}_{\mathrm{MF}}
=\operatorname{AdaptiveL2}(
\mathbf{u}_{\theta}-\mathbf{u}_{\mathrm{tgt}})$.

\rev{In SOAR, $u_\theta$ is induced by the calibrated action through
Eq.~(6), rather than predicted by an independent velocity head.
The underlying data-to-average-velocity conversion has the same
algebraic form as pixel MeanFlow~\citep{lu2026pixelmeanflow},
away from the numerical clamp.
SOAR adapts this formulation to executable action chunks and
integrates it with an observation-conditioned trajectory-structure
predictor.
The predicted structure conditions the generator and calibrates
its output before the MeanFlow objective is evaluated.
Consequently, $\mathcal{L}_{\mathrm{MF}}$ constrains the
structurally guided action produced by the policy.
Together with the inference-endpoint objective in Appendix A.1.3,
this design couples predicted temporal structure with both
one-step generation and training of the resulting action,
while retaining a single evaluation of the generative backbone
at inference.}

\noindent\textbf{Adaptive regression.}
For a per-sample residual $\boldsymbol{\delta}$, the AdaptiveL2 metric is
implemented as
\begin{equation}
\begin{aligned}
    \operatorname{AdaptiveL2}(\boldsymbol{\delta})
    &=
    \frac{1}{B}
    \sum_{b=1}^{B}
    \operatorname{sg}
    \left[
        (m_b+10^{-3})^{-1/2}
    \right]m_b,\\
    m_b
    &=
    \operatorname{mean}
    \left(
        \boldsymbol{\delta}_b^2
    \right).
\end{aligned}
\label{eq:app_adaptive_l2}
\end{equation}
The detached factor adaptively rescales sample-wise regression errors while
preventing gradients from modifying the weighting term itself.


\subsubsection{Inference-Endpoint Alignment Details}
\label{app:endpoint_details}

Random-interval MeanFlow training does not explicitly optimize the fixed
condition encountered during one-step inference. SOAR therefore introduces
an additional objective at $(t,r)=(1,0)$.

We independently sample
$\boldsymbol{\epsilon}_{\mathrm{end}}
\sim\mathcal{N}(\mathbf{0},\mathbf{I})$
and evaluate the same action generator under this inference condition.
The corresponding ground-truth and model-induced endpoint velocities are
$\mathbf{u}^{\mathrm{gt}}_{\mathrm{end}}
=\boldsymbol{\epsilon}_{\mathrm{end}}-\mathbf{x}$,
$\mathbf{u}^{\theta}_{\mathrm{end}}
=\boldsymbol{\epsilon}_{\mathrm{end}}
-\widehat{\mathbf{A}}_{\mathrm{end}}$,
where $\widehat{\mathbf{A}}_{\mathrm{end}}$ denotes the calibrated action
predicted at $(t,r)=(1,0)$.

The velocity-space endpoint objective is
\begin{equation}
    \mathcal{L}_{\mathrm{end}}
    =
    \operatorname{AdaptiveL2}
    \left(
        \mathbf{u}^{\theta}_{\mathrm{end}}
        -
        \mathbf{u}^{\mathrm{gt}}_{\mathrm{end}}
    \right).
    \label{eq:app_endpoint_velocity_loss}
\end{equation}

At $t=1$, the two endpoint velocities share the same noise sample, yielding
$\mathbf{u}^{\theta}_{\mathrm{end}}
-\mathbf{u}^{\mathrm{gt}}_{\mathrm{end}}
=\mathbf{x}-\widehat{\mathbf{A}}_{\mathrm{end}}$.
Therefore, ~\eqref{eq:app_endpoint_velocity_loss} is algebraically equivalent to
$\mathcal{L}_{\mathrm{end}}
=\operatorname{AdaptiveL2}(
\mathbf{x}-\widehat{\mathbf{A}}_{\mathrm{end}})$,
which is the compact form used in the main text.
The velocity-space formulation emphasizes that the objective aligns the
actual MeanFlow sampling boundary rather than the robot's physical dynamics.


\subsubsection{Dispersive Feature Regularization}
\label{app:dispersive_details}

Following MP1~\citep{sheng2026mp1}, we retain a dispersive regularizer on
intermediate representations of the action U-Net.
For hidden representation $\mathbf{h}^{(l)}$ at layer $l$, we compute the
pairwise squared feature distance
$d_{ij}^{(l)}
=\|\mathbf{h}^{(l)}_i-\mathbf{h}^{(l)}_j\|_2^2$.

Using normalized distances $\bar d_{ij}^{(l)}$, the regularizer is
\begin{equation}
    \mathcal{L}_{\mathrm{dis}}
    =
    \sum_l
    \log
    \left[
        \frac{1}{B^2}
        \sum_{i,j}
        \exp
        \left(
            -
            \frac{
                \bar d_{ij}^{(l)}
            }{\tau}
        \right)
    \right],
    \qquad
    \tau=1.
    \label{eq:app_dispersive_loss}
\end{equation}
Minimizing this objective encourages different samples to maintain
discriminative hidden representations and helps reduce representation
collapse or overly averaged predictions in one-step generation.


\rev{\subsubsection{Learned Calibration and Conditioning Parameters}
\label{app:learned_parameters}

\paragraph{Configuration.}
We inspect the best EMA checkpoints for all three Adroit tasks and
all 50 Meta-World tasks, with three training seeds per task.
This yields 9 Adroit checkpoints and 150 Meta-World checkpoints.
All inspected checkpoints use the complete DCT representation with
$K=T=4$. The low-, mid-, and high-frequency groups contain
coefficient indices $\{0,1\}$, $\{2\}$, and $\{3\}$, respectively.
The conditioning gates $g_L$, $g_M$, and $g_H$ in Eq.~(8) are
unconstrained learnable scalars applied directly to the projected
features, without sigmoid or tanh activation.

The fixed loss weights in Eq.~(1) are
$(\lambda_{\mathrm{struct}},\lambda_{\mathrm{act}},
\lambda_{\mathrm{dis}},\lambda_{\mathrm{end}})
=(0.05,0.01,0.5,0.05)$,
and the coefficient of $\mathcal{L}_{\mathrm{MF}}$ is 1. Across benchmarks, we share the outer loss coefficients in Equation (1), the orthonormal temporal DCT-II convention, the full-spectrum representation, and the structure-head hidden width of 512. Structure-head depth and activation, observation encoders, prediction and execution lengths, initialization, and optimization schedules are setting-specific. The additional temporal regularization, gripper weighting, event sampling, and RTC probabilities reported in Appendix A.4 describe real-robot adaptation and are not assumed to be shared across simulation benchmarks.

\paragraph{Learned parameter statistics.}
Table~\ref{tab:learned_parameters} reports the learned calibration
coefficient and conditioning gates. Means and sample standard
deviations are computed across task--seed checkpoints.
These statistics describe variation across trained models,
rather than within-task training-seed uncertainty.

\begin{table}[tbp]
    \centering
    \caption{
        Learned calibration and conditioning parameters across
        9 Adroit checkpoints and 150 Meta-World checkpoints.
        Statistics cover all three Adroit tasks and all 50 Meta-World
        tasks, with three training seeds per task.
        Values are mean $\pm$ sample standard deviation across
        task--seed checkpoints, except for the range of $\alpha$
        and the mean covariance scaling factor.
        Here, $\alpha := \tanh(\rho)$.
    }
    \label{tab:learned_parameters}
    \vspace{2mm}

    \setlength{\tabcolsep}{5pt}
    \renewcommand{\arraystretch}{1.12}

    \resizebox{\textwidth}{!}{%
    \begin{tabular}{lcccccc}
        \toprule
        \textbf{Benchmark}
        & $\boldsymbol{\alpha}$
        & \textbf{Range of $\boldsymbol{\alpha}$}
        & $\boldsymbol{g_L}$
        & $\boldsymbol{g_M}$
        & $\boldsymbol{g_H}$
        & \textbf{Mean $(1-\boldsymbol{\alpha})^2$} \\
        \midrule

        Adroit
        & $0.118051 \pm 0.081276$
        & $[0.007725,\,0.220070]$
        & $-0.002653 \pm 0.004563$
        & $0.000278 \pm 0.005642$
        & $0.001601 \pm 0.003492$
        & $0.7837$ \\

        Meta-World
        & $0.075216 \pm 0.053296$
        & $[0.015644,\,0.273769]$
        & $0.000244 \pm 0.009540$
        & $-0.001454 \pm 0.004646$
        & $-0.003051 \pm 0.005112$
        & $0.8580$ \\

        \bottomrule
    \end{tabular}%
    }
\end{table}

\paragraph{Relation to deterministic regression.}
For the complete DCT used here, linearity and invertibility make
Eq.~(4) exactly equivalent to
\begin{equation}
    \hat{A}
    = (1-\alpha)A_{\mathrm{base}}
    + \alpha\,\operatorname{IDCT}(C^s(o)),
    \qquad \alpha := \tanh(\rho).
    \label{eq:calibration_affine}
\end{equation}
Since $C^s(o)$ depends only on the observation, the output approaches
the deterministic structure prediction as $\alpha$ approaches 1.
However, the measured coefficients are far from this limit:
the maximum values are 0.220070 on Adroit and 0.273769 on Meta-World.

For a fixed observation and checkpoint, randomness arises from
the Gaussian latent $\epsilon$, giving
\begin{equation}
    \operatorname{Cov}_{\epsilon}[\hat{A}\mid o]
    = (1-\alpha)^2
      \operatorname{Cov}_{\epsilon}[A_{\mathrm{base}}\mid o].
    \label{eq:calibration_covariance}
\end{equation}
The checkpoint-averaged covariance scaling factors are 0.7837
on Adroit and 0.8580 on Meta-World.
These results indicate that SOAR preserves stochastic
action generation rather than approaching deterministic
regression after calibration.

}

\subsection{More Details on Simulation Experiment With 3D Inputs}
\label{sec:3D}

\subsubsection{More Details on Inference Latency Comparison}
\label{sec:latency_details}

Table~\ref{tab:latency} reports the detailed inference latency on
Adroit and Meta-World. Inference latency is measured as the wall-clock
time required for one complete policy evaluation to generate an action
chunk. All results are reported as mean $\pm$ standard deviation over
three random seeds. For FlowPolicy, three repeated measurements are
averaged for each seed before computing the final statistics. The latency increases with the number of network evaluations used for
action generation. DP requires $678.4 \pm 6.9$\,ms with
$\mathrm{NFE}=100$, while DP3 and Simple DP3 require
$70.7 \pm 2.6$\,ms and $43.4 \pm 0.4$\,ms, respectively, with
$\mathrm{NFE}=10$. Among the one-step methods, FlowPolicy requires
$14.7 \pm 0.0$\,ms, MP1 requires $8.1 \pm 0.0$\,ms, and SOAR requires
$8.7 \pm 0.1$\,ms. SOAR maintains similar latency across different tasks, ranging from
$8.6$ to $8.9$\,ms on Adroit and Meta-World. Compared with MP1, the
additional structure prediction and action calibration introduce only
a small latency increase, while the generative backbone is still
evaluated once. These measurements confirm that SOAR retains
millisecond-level inference with $\mathrm{NFE}=1$.

\subsubsection{More Details on Performance}
\label{sec:3d_taskwise}

Table~\ref{tab:53_taskwise_results} provides the task-wise results on the
53-task benchmark. Each result is reported as mean success rate $\pm$
sample standard deviation over three random seeds. Most Easy tasks are close to saturation for several methods, while the
differences become more evident on Medium, Hard, and Very Hard tasks.
SOAR performs strongly on a range of non-saturated tasks, including
Peg Insert Side, Push Wall, Sweep Into, Pick Out Of Hole, Shelf Place,
Stick Pull, and Pick Place Wall. It also improves several Adroit and Easy
tasks such as Door, Handle Pull, Handle Pull Side, and Lever Pull.
Performance remains task-dependent, with other methods retaining an
advantage on several tasks. The detailed results therefore indicate that
the overall gain of SOAR comes mainly from broader improvements on
non-saturated tasks.

\begin{table}[tbp]
\centering
\caption{
Inference latency comparison on the Adroit and Meta-World benchmarks.
Inference latency measures the wall-clock time of one complete policy
evaluation that generates an action chunk.
Results are reported as mean latency (ms) $\pm$ standard deviation over
three random seeds.
For FlowPolicy, three repeated measurements are averaged for each seed.
\textbf{Bold} denotes the best result.
}
\label{tab:latency}

\setlength{\tabcolsep}{4.2pt}
\renewcommand{\arraystretch}{1.10}

\resizebox{\textwidth}{!}{%
\begin{tabular}{lccccccccccc}
    \toprule

    \textbf{}
    & \textbf{}
    & \textbf{}
    & \multicolumn{3}{c}{\textbf{Adroit}}
    & \multicolumn{4}{c}{\textbf{Meta-World}}
    & \textbf{} \\

    \cmidrule(lr){4-6}
    \cmidrule(lr){7-10}

    \textbf{Methods}
    & \textbf{Publication}
    & \textbf{NFE}
    & \textbf{Hammer}
    & \textbf{Door}
    & \textbf{Pen}
    & \textbf{Easy (28)}
    & \textbf{Medium (11)}
    & \textbf{Hard (6)}
    & \textbf{Very Hard (5)}
    & \textbf{Inference Latency (ms)} \\

    \midrule

    DP~\citep{chi2023diffusion}
    & RSS, 2023
    & 100
    & $671.6 \pm 8.7$
    & $693.9 \pm 16.9$
    & $676.4 \pm 2.7$
    & $676.5 \pm 6.8$
    & $672.7 \pm 5.8$
    & $681.0 \pm 9.2$
    & $696.7 \pm 23.4$
    & $678.4 \pm 6.9$ \\

    DP3~\citep{ze2024dp3}
    & RSS, 2024
    & 10
    & $69.0 \pm 0.9$
    & $70.4 \pm 1.8$
    & $80.6 \pm 3.4$
    & $69.3 \pm 1.7$
    & $69.9 \pm 1.2$
    & $72.7 \pm 2.3$
    & $76.5 \pm 1.6$
    & $70.7 \pm 2.6$ \\

    Simple DP3~\citep{ze2024dp3}
    & RSS, 2024
    & 10
    & $43.7 \pm 0.3$
    & $43.3 \pm 0.1$
    & $43.6 \pm 0.4$
    & $43.3 \pm 0.4$
    & $43.4 \pm 0.4$
    & $43.3 \pm 0.4$
    & $43.4 \pm 0.3$
    & $43.4 \pm 0.4$ \\

    FlowPolicy~\citep{zhang2025flowpolicy}
    & AAAI, 2025
    & 1
    & $14.6 \pm 0.1$
    & $14.7 \pm 0.1$
    & $14.6 \pm 0.0$
    & $14.7 \pm 0.0$
    & $14.8 \pm 0.0$
    & $14.7 \pm 0.0$
    & $14.7 \pm 0.1$
    & $14.7 \pm 0.0$ \\

    MP1~\citep{sheng2026mp1}
    & AAAI, 2026
    & 1
    & $\mathbf{8.2 \pm 0.0}$
    & $\mathbf{8.2 \pm 0.1}$
    & $\mathbf{8.2 \pm 0.0}$
    & $\mathbf{8.1 \pm 0.1}$
    & $\mathbf{8.1 \pm 0.1}$
    & $\mathbf{8.0 \pm 0.1}$
    & $\mathbf{8.1 \pm 0.1}$
    & $\mathbf{8.1 \pm 0.0}$ \\

    \midrule

    \textbf{SOAR (Ours)}
    & \textbf{--}
    & \textbf{1}
    & $8.7 \pm 0.3$
    & $8.6 \pm 0.6$
    & $8.6 \pm 0.4$
    & $8.7 \pm 0.2$
    & $8.6 \pm 0.0$
    & $8.6 \pm 0.0$
    & $8.9 \pm 0.1$
    & $8.7 \pm 0.1$ \\

    \bottomrule
\end{tabular}%
}
\end{table}

\begin{table}[p]
    \centering
    \caption{Detailed task-wise performance on the 53-task benchmark.
        Results are reported as mean success rate (\%) $\pm$ sample standard
        deviation over three random seeds.
        \textbf{Bold} denotes the best result in each column.}
    \label{tab:53_taskwise_results}

    \setlength{\tabcolsep}{2.1pt}
    \renewcommand{\arraystretch}{1.22}
    \fontsize{10}{10}\selectfont
    \setlength{\aboverulesep}{0.15ex}
    \setlength{\belowrulesep}{0.25ex}
    \setlength{\abovetopsep}{0pt}
    \setlength{\belowbottomsep}{0pt}
    \setlength{\parskip}{0pt}
    \setlength{\lineskip}{0pt}

    \resizebox{\linewidth}{!}{%
    \begin{tabular}{lcccccccc}
        \toprule
        & & & \multicolumn{3}{c}{\textbf{Adroit}}
              & \multicolumn{3}{c}{\textbf{Meta-World (Easy)}} \\
        \cmidrule(lr){4-6}
        \cmidrule(lr){7-9}
        \textbf{Methods} & \textbf{Publication} & \textbf{NFE}
        & \textbf{Hammer} & \textbf{Door} & \textbf{Pen}
        & \textbf{Button Press} & \textbf{Coffee Button} & \textbf{Plate Slide Back Side} \\
        \midrule
        DP & RSS, 2023 & 100 & $75.1 \pm 3.5$ & $23.7 \pm 3.7$ & $19.3 \pm 1.8$ & $79.7 \pm 3.8$ & $83.0 \pm 3.5$ & $99.3 \pm 0.6$ \\
        DP3 & RSS, 2024 & 10 & $\mathbf{100.0 \pm 0.0}$ & $73.5 \pm 2.1$ & $56.7 \pm 2.6$ & $\mathbf{100.0 \pm 0.0}$ & $\mathbf{100.0 \pm 0.0}$ & $\mathbf{100.0 \pm 0.0}$ \\
        Simple DP3 & RSS, 2024 & 10 & $\mathbf{100.0 \pm 0.0}$ & $56.8 \pm 1.2$ & $56.1 \pm 0.3$ & $\mathbf{100.0 \pm 0.0}$ & $\mathbf{100.0 \pm 0.0}$ & $\mathbf{100.0 \pm 0.0}$ \\
        FlowPolicy & AAAI, 2025 & 1 & $95.2 \pm 2.4$ & $57.3 \pm 6.2$ & $47.8 \pm 11.0$ & $\mathbf{100.0 \pm 0.0}$ & $\mathbf{100.0 \pm 0.0}$ & $\mathbf{100.0 \pm 0.0}$ \\
        MP1 & AAAI, 2026 & 1 & $96.1 \pm 2.3$ & $69.2 \pm 3.1$ & $57.1 \pm 6.1$ & $\mathbf{100.0 \pm 0.0}$ & $\mathbf{100.0 \pm 0.0}$ & $\mathbf{100.0 \pm 0.0}$ \\
        \midrule
        \textbf{SOAR (Ours)} & - & 1 & $\mathbf{100.0 \pm 0.0}$ & $\mathbf{77.5 \pm 2.4}$ & $\mathbf{61.9 \pm 7.3}$ & $\mathbf{100.0 \pm 0.0}$ & $\mathbf{100.0 \pm 0.0}$ & $\mathbf{100.0 \pm 0.0}$ \\
        \bottomrule
    \end{tabular}%
    }

    \vspace{0.40em}

    \resizebox{\linewidth}{!}{%
    \begin{tabular}{lcccccc}
        \toprule
        & \multicolumn{6}{c}{\textbf{Meta-World (Easy)}} \\
        \cmidrule(lr){2-7}
        \textbf{Methods} & \textbf{Button Press Topdown}
        & \textbf{Button Press Topdown Wall} & \textbf{Button Press Wall}
        & \textbf{Peg Unplug Side} & \textbf{Door Close} & \textbf{Door Lock} \\
        \midrule
        DP & $42.7 \pm 7.6$ & $39.3 \pm 5.7$ & $87.7 \pm 5.7$ & $32.0 \pm 12.3$ & $\mathbf{100.0 \pm 0.0}$ & $66.3 \pm 5.9$ \\
        DP3 & $96.7 \pm 5.8$ & $88.0 \pm 6.2$ & $\mathbf{100.0 \pm 0.0}$ & $71.3 \pm 4.7$ & $\mathbf{100.0 \pm 0.0}$ & $73.7 \pm 12.1$ \\
        Simple DP3 & $96.0 \pm 6.1$ & $87.3 \pm 7.8$ & $\mathbf{100.0 \pm 0.0}$ & $65.3 \pm 14.2$ & $\mathbf{100.0 \pm 0.0}$ & $74.3 \pm 21.1$ \\
        FlowPolicy & $\mathbf{100.0 \pm 0.0}$ & $\mathbf{100.0 \pm 0.0}$ & $\mathbf{100.0 \pm 0.0}$ & $\mathbf{85.0 \pm 2.0}$ & $\mathbf{100.0 \pm 0.0}$ & $\mathbf{100.0 \pm 0.0}$ \\
        MP1 & $\mathbf{100.0 \pm 0.0}$ & $\mathbf{100.0 \pm 0.0}$ & $99.7 \pm 0.6$ & $67.0 \pm 4.6$ & $\mathbf{100.0 \pm 0.0}$ & $97.3 \pm 2.9$ \\
        \midrule
        \textbf{SOAR (Ours)} & $\mathbf{100.0 \pm 0.0}$ & $\mathbf{100.0 \pm 0.0}$ & $\mathbf{100.0 \pm 0.0}$ & $83.7 \pm 6.8$ & $\mathbf{100.0 \pm 0.0}$ & $\mathbf{100.0 \pm 0.0}$ \\
        \bottomrule
    \end{tabular}%
    }

    \vspace{0.40em}

    \resizebox{\linewidth}{!}{%
    \begin{tabular}{lcccccccc}
        \toprule
        & \multicolumn{8}{c}{\textbf{Meta-World (Easy)}} \\
        \cmidrule(lr){2-9}
        \textbf{Methods} & \textbf{Door Open} & \textbf{Door Unlock}
        & \textbf{Drawer Close} & \textbf{Drawer Open} & \textbf{Faucet Close}
        & \textbf{Faucet Open} & \textbf{Handle Press} & \textbf{Handle Pull} \\
        \midrule
        DP & $93.3 \pm 2.5$ & $98.7 \pm 1.2$ & $97.7 \pm 2.1$ & $55.7 \pm 9.5$ & $98.3 \pm 1.5$ & $89.0 \pm 3.0$ & $77.7 \pm 2.5$ & $9.0 \pm 7.0$ \\
        DP3 & $81.7 \pm 3.5$ & $99.3 \pm 1.2$ & $\mathbf{100.0 \pm 0.0}$ & $97.0 \pm 1.7$ & $78.3 \pm 11.4$ & $\mathbf{100.0 \pm 0.0}$ & $84.0 \pm 1.0$ & $21.3 \pm 3.2$ \\
        Simple DP3 & $86.7 \pm 6.4$ & $99.3 \pm 1.2$ & $\mathbf{100.0 \pm 0.0}$ & $96.3 \pm 4.0$ & $90.7 \pm 8.1$ & $\mathbf{100.0 \pm 0.0}$ & $84.0 \pm 6.6$ & $11.0 \pm 5.3$ \\
        FlowPolicy & $\mathbf{100.0 \pm 0.0}$ & $\mathbf{100.0 \pm 0.0}$ & $\mathbf{100.0 \pm 0.0}$ & $\mathbf{100.0 \pm 0.0}$ & $\mathbf{100.0 \pm 0.0}$ & $\mathbf{100.0 \pm 0.0}$ & $98.3 \pm 2.9$ & $13.7 \pm 3.2$ \\
        MP1 & $96.7 \pm 0.6$ & $98.7 \pm 2.3$ & $\mathbf{100.0 \pm 0.0}$ & $\mathbf{100.0 \pm 0.0}$ & $\mathbf{100.0 \pm 0.0}$ & $\mathbf{100.0 \pm 0.0}$ & $99.7 \pm 0.6$ & $41.3 \pm 12.4$ \\
        \midrule
        \textbf{SOAR (Ours)} & $\mathbf{100.0 \pm 0.0}$ & $\mathbf{100.0 \pm 0.0}$ & $\mathbf{100.0 \pm 0.0}$ & $\mathbf{100.0 \pm 0.0}$ & $\mathbf{100.0 \pm 0.0}$ & $\mathbf{100.0 \pm 0.0}$ & $\mathbf{100.0 \pm 0.0}$ & $\mathbf{67.7 \pm 15.0}$ \\
        \bottomrule
    \end{tabular}%
    }

    \vspace{0.40em}

    \resizebox{\linewidth}{!}{%
    \begin{tabular}{lcccccccc}
        \toprule
        & \multicolumn{8}{c}{\textbf{Meta-World (Easy)}} \\
        \cmidrule(lr){2-9}
        \textbf{Methods} & \textbf{Handle Press Side} & \textbf{Handle Pull Side}
        & \textbf{Lever Pull} & \textbf{Plate Slide} & \textbf{Plate Slide Back}
        & \textbf{Dial Turn} & \textbf{Reach} & \textbf{Reach Wall} \\
        \midrule
        DP & $60.3 \pm 6.1$ & $6.3 \pm 1.2$ & $44.0 \pm 8.7$ & $91.0 \pm 1.0$ & $\mathbf{100.0 \pm 0.0}$ & $23.3 \pm 3.8$ & $37.7 \pm 5.7$ & $31.0 \pm 10.5$ \\
        DP3 & $48.3 \pm 10.8$ & $50.3 \pm 9.9$ & $77.3 \pm 5.5$ & $\mathbf{100.0 \pm 0.0}$ & $\mathbf{100.0 \pm 0.0}$ & $55.3 \pm 5.5$ & $44.0 \pm 9.5$ & $67.3 \pm 3.1$ \\
        Simple DP3 & $45.0 \pm 7.0$ & $35.7 \pm 11.0$ & $72.3 \pm 12.9$ & $\mathbf{100.0 \pm 0.0}$ & $\mathbf{100.0 \pm 0.0}$ & $56.0 \pm 2.0$ & $41.3 \pm 11.0$ & $59.7 \pm 8.1$ \\
        FlowPolicy & $\mathbf{100.0 \pm 0.0}$ & $54.0 \pm 5.6$ & $79.3 \pm 3.5$ & $\mathbf{100.0 \pm 0.0}$ & $\mathbf{100.0 \pm 0.0}$ & $85.3 \pm 9.1$ & $\mathbf{45.3 \pm 5.1}$ & $\mathbf{76.3 \pm 3.1}$ \\
        MP1 & $\mathbf{100.0 \pm 0.0}$ & $56.7 \pm 7.1$ & $79.7 \pm 2.1$ & $\mathbf{100.0 \pm 0.0}$ & $97.7 \pm 1.5$ & $70.0 \pm 8.2$ & $14.3 \pm 2.1$ & $62.0 \pm 7.0$ \\
        \midrule
        \textbf{SOAR (Ours)} & $\mathbf{100.0 \pm 0.0}$ & $\mathbf{84.7 \pm 8.5}$ & $\mathbf{96.0 \pm 1.0}$ & $\mathbf{100.0 \pm 0.0}$ & $\mathbf{100.0 \pm 0.0}$ & $\mathbf{88.3 \pm 1.5}$ & $27.7 \pm 4.0$ & $74.7 \pm 2.5$ \\
        \bottomrule
    \end{tabular}%
    }

    \vspace{0.40em}

    \resizebox{\linewidth}{!}{%
    \begin{tabular}{lcccccccc}
        \toprule
        & \multicolumn{3}{c}{\textbf{Meta-World (Easy)}}
        & \multicolumn{5}{c}{\textbf{Meta-World (Medium)}} \\
        \cmidrule(lr){2-4}
        \cmidrule(lr){5-9}
        \textbf{Methods} & \textbf{Plate Slide Side} & \textbf{Window Close}
        & \textbf{Window Open} & \textbf{Basketball} & \textbf{Bin Picking}
        & \textbf{Box Close} & \textbf{Coffee Pull} & \textbf{Coffee Push} \\
        \midrule
        DP & $98.0 \pm 2.0$ & $99.0 \pm 1.7$ & $92.0 \pm 5.3$ & $80.7 \pm 2.5$ & $20.3 \pm 0.6$ & $36.7 \pm 5.7$ & $22.0 \pm 5.3$ & $54.3 \pm 2.1$ \\
        DP3 & $\mathbf{100.0 \pm 0.0}$ & $\mathbf{100.0 \pm 0.0}$ & $94.0 \pm 4.4$ & $84.0 \pm 5.0$ & $23.3 \pm 5.1$ & $50.3 \pm 11.0$ & $58.0 \pm 25.1$ & $\mathbf{74.0 \pm 5.6}$ \\
        Simple DP3 & $\mathbf{100.0 \pm 0.0}$ & $\mathbf{100.0 \pm 0.0}$ & $93.7 \pm 5.7$ & $66.3 \pm 4.9$ & $21.3 \pm 6.1$ & $46.3 \pm 8.1$ & $53.0 \pm 18.1$ & $69.3 \pm 3.2$ \\
        FlowPolicy & $\mathbf{100.0 \pm 0.0}$ & $\mathbf{100.0 \pm 0.0}$ & $\mathbf{100.0 \pm 0.0}$ & $97.7 \pm 2.1$ & $22.7 \pm 4.7$ & $49.7 \pm 9.5$ & $67.7 \pm 4.0$ & $64.0 \pm 6.6$ \\
        MP1 & $\mathbf{100.0 \pm 0.0}$ & $\mathbf{100.0 \pm 0.0}$ & $98.7 \pm 2.3$ & $80.0 \pm 4.4$ & $15.3 \pm 9.1$ & $49.3 \pm 11.2$ & $82.0 \pm 6.0$ & $59.3 \pm 3.1$ \\
        \midrule
        \textbf{SOAR (Ours)} & $\mathbf{100.0 \pm 0.0}$ & $\mathbf{100.0 \pm 0.0}$ & $\mathbf{100.0 \pm 0.0}$ & $\mathbf{100.0 \pm 0.0}$ & $\mathbf{34.0 \pm 7.5}$ & $\mathbf{65.7 \pm 5.5}$ & $\mathbf{96.0 \pm 2.6}$ & $70.7 \pm 3.2$ \\
        \bottomrule
    \end{tabular}%
    }

    \vspace{0.40em}

    \resizebox{\linewidth}{!}{%
    \begin{tabular}{lccccccccc}
        \toprule
        & \multicolumn{6}{c}{\textbf{Meta-World (Medium)}}
        & \multicolumn{3}{c}{\textbf{Meta-World (Hard)}} \\
        \cmidrule(lr){2-7}
        \cmidrule(lr){8-10}
        \textbf{Methods} & \textbf{Hammer} & \textbf{Peg Insert Side}
        & \textbf{Push Wall} & \textbf{Soccer} & \textbf{Sweep} & \textbf{Sweep Into}
        & \textbf{Assembly} & \textbf{Hand Insert} & \textbf{Pick Out Of Hole} \\
        \midrule
        DP & $36.7 \pm 15.3$ & $31.0 \pm 13.7$ & $25.0 \pm 15.6$ & $8.0 \pm 4.6$ & $32.7 \pm 10.6$ & $13.0 \pm 3.0$ & $71.3 \pm 7.0$ & $10.3 \pm 3.1$ & $27.3 \pm 8.6$ \\
        DP3 & $93.7 \pm 9.3$ & $62.7 \pm 4.5$ & $74.3 \pm 6.7$ & $11.7 \pm 5.0$ & $98.7 \pm 1.5$ & $14.7 \pm 4.2$ & $94.0 \pm 2.0$ & $9.7 \pm 2.3$ & $15.7 \pm 4.5$ \\
        Simple DP3 & $89.3 \pm 9.7$ & $61.7 \pm 11.2$ & $58.3 \pm 11.0$ & $14.3 \pm 4.9$ & $91.0 \pm 7.9$ & $20.3 \pm 4.5$ & $93.7 \pm 3.8$ & $8.3 \pm 0.6$ & $14.7 \pm 12.9$ \\
        FlowPolicy & $78.7 \pm 6.0$ & $63.7 \pm 4.6$ & $36.3 \pm 7.0$ & $14.7 \pm 6.4$ & $99.7 \pm 0.6$ & $21.0 \pm 7.5$ & $96.7 \pm 5.8$ & $9.7 \pm 3.5$ & $26.3 \pm 14.6$ \\
        MP1 & $98.3 \pm 1.2$ & $75.7 \pm 3.5$ & $72.7 \pm 5.0$ & $13.0 \pm 5.6$ & $93.7 \pm 2.5$ & $15.0 \pm 1.7$ & $98.0 \pm 2.6$ & $9.0 \pm 2.6$ & $11.7 \pm 6.8$ \\
        \midrule
        \textbf{SOAR (Ours)} & $\mathbf{100.0 \pm 0.0}$ & $\mathbf{88.7 \pm 4.0}$ & $\mathbf{99.7 \pm 0.6}$ & $\mathbf{22.7 \pm 6.0}$ & $\mathbf{100.0 \pm 0.0}$ & $\mathbf{39.7 \pm 12.9}$ & $\mathbf{100.0 \pm 0.0}$ & $\mathbf{13.0 \pm 4.0}$ & $\mathbf{42.3 \pm 1.5}$ \\
        \bottomrule
    \end{tabular}%
    }

    \vspace{0.40em}

    \resizebox{\linewidth}{!}{%
    \begin{tabular}{lccccccccc}
        \toprule
        & \multicolumn{3}{c}{\textbf{Meta-World (Hard)}}
        & \multicolumn{5}{c}{\textbf{Meta-World (Very Hard)}} & \textbf{Overall} \\
        \cmidrule(lr){2-4}
        \cmidrule(lr){5-9}
        \cmidrule(lr){10-10}
        \textbf{Methods} & \textbf{Pick Place} & \textbf{Push} & \textbf{Push Back}
        & \textbf{Shelf Place} & \textbf{Disassemble} & \textbf{Stick Pull}
        & \textbf{Stick Push} & \textbf{Pick Place Wall} & \textbf{Average} \\
        \midrule
        DP & $17.7 \pm 8.1$ & $24.3 \pm 8.5$ & $54.7 \pm 13.1$ & $4.3 \pm 0.6$ & $62.3 \pm 3.5$ & $21.0 \pm 2.5$ & $53.7 \pm 10.8$ & $16.3 \pm 12.9$ & $52.3 \pm 0.9$ \\
        DP3 & $58.3 \pm 3.5$ & $72.0 \pm 13.5$ & $74.7 \pm 12.7$ & $15.7 \pm 4.7$ & $63.7 \pm 6.1$ & $45.0 \pm 8.2$ & $96.0 \pm 4.4$ & $83.3 \pm 3.1$ & $72.3 \pm 1.4$ \\
        Simple DP3 & $60.0 \pm 3.6$ & $73.3 \pm 15.3$ & $75.7 \pm 7.8$ & $23.0 \pm 7.2$ & $59.0 \pm 4.4$ & $49.3 \pm 14.6$ & $96.7 \pm 4.2$ & $80.7 \pm 5.5$ & $70.4 \pm 1.5$ \\
        FlowPolicy & $50.7 \pm 12.9$ & $58.0 \pm 25.6$ & $\mathbf{85.0 \pm 13.2}$ & $17.3 \pm 3.8$ & $\mathbf{87.7 \pm 8.5}$ & $45.3 \pm 11.5$ & $98.7 \pm 2.3$ & $64.3 \pm 3.1$ & $75.3 \pm 1.8$ \\
        MP1 & $60.7 \pm 8.4$ & $74.7 \pm 11.5$ & $82.7 \pm 1.2$ & $24.0 \pm 11.4$ & $62.7 \pm 5.0$ & $68.7 \pm 1.5$ & $95.7 \pm 7.5$ & $77.3 \pm 3.1$ & $75.9 \pm 1.3$ \\
        \midrule
        \textbf{SOAR (Ours)} & $\mathbf{64.3 \pm 8.1}$ & $\mathbf{76.7 \pm 4.7}$ & $84.7 \pm 4.7$ & $\mathbf{43.3 \pm 13.3}$ & $79.7 \pm 6.5$ & $\mathbf{73.0 \pm 2.0}$ & $\mathbf{100.0 \pm 0.0}$ & $\mathbf{88.7 \pm 3.1}$ & $\mathbf{83.9 \pm 0.3}$ \\
        \bottomrule
    \end{tabular}%
    }
\end{table}

\subsection{Additional VLA Experimental Details}
\label{sec:vla_details}

This section provides additional implementation, training, and evaluation
details for the VLA experiments in Section~\ref{sec:vla_simulation}.
In both benchmarks, SOAR replaces the iterative action expert of
$\pi_{0.5}$ while retaining the original vision-language backbone and
data-processing pipeline.

\subsubsection{LIBERO}
\label{sec:libero_details}

We evaluate $\pi_{0.5}$-SOAR on all four standard LIBERO
suites~\citep{liu2023libero}: LIBERO-Spatial, LIBERO-Object,
LIBERO-Goal, and LIBERO-Long. Each suite contains 10 tasks, resulting in
40 tasks in total. A single policy is jointly trained across all tasks. The vision-language backbone is initialized from the OpenPI
\texttt{pi05\_base} checkpoint. The original iterative action expert is
replaced by SOAR, and the newly introduced SOAR parameters are initialized
from scratch. We fine-tune the model for 30K optimization steps using
AdamW with a global batch size of 256 on four NVIDIA A100-80GB GPUs. SOAR predicts a 10-step action chunk at each policy query. During
evaluation, the first five predicted actions are executed before the
policy receives a new observation and replans. Unlike the original
$\pi_{0.5}$ action expert, which uses multiple network evaluations for
action generation, SOAR generates the complete action chunk with a single
action-expert evaluation ($\mathrm{NFE}=1$). Each task is evaluated for
50 episodes.

\subsubsection{RoboTwin 2.0}
\label{sec:robotwin_details}

\rev{

We further evaluate $\pi_{0.5}$-SOAR on RoboTwin
2.0~\citep{chen2025robotwin}. We jointly train a single
multi-task policy on the clean demonstrations from all 50 tasks.
The clean-only training subset contains 2,500 trajectories
(50 per task) and 549,787 frames. We use task-balanced sampling,
with uniform frame sampling within each task and sampling
with replacement. Each observation comprises three RGB views, including one external
camera and two wrist cameras, together with a 14-dimensional
proprioceptive state. RGB observations are processed at
$224\times224$ resolution. The vision-language backbone is
initialized from the OpenPI \texttt{pi05\_base} checkpoint.
The original flow-based action expert is replaced by the
SOAR action module, which is initialized using pretrained
weights. We fine-tune all model parameters for 100K optimization steps
on four NVIDIA A100-80GB GPUs using FSDP. The global microbatch
size is 64, with two gradient-accumulation steps, yielding an
effective batch size of 128. We use AdamW with
$\beta_1=0.9$, $\beta_2=0.95$, $\epsilon=10^{-8}$,
weight decay 0.01, and gradient-norm clipping at 1.0.
The learning rate increases to $2.5\times10^{-5}$ over
1K warmup steps, then follows a cosine schedule to
$2.5\times10^{-6}$ at step 60K and remains constant
until step 100K. An exponential moving average with
decay 0.999 is maintained. States and actions are
normalized using statistics computed from the
clean-only training subset.

The policy predicts 50-step chunks of 14-dimensional absolute
joint-position and gripper actions. State and action vectors
are zero-padded to 32 dimensions, while the action loss is
computed only over the 14 valid dimensions.
At each policy invocation, SOAR generates the complete action
chunk with a single action-expert evaluation
($\mathrm{NFE}=1$). All 50 actions are executed open-loop
before replanning. We evaluate the final checkpoint on all
50 tasks using 100 episodes per task.

}

\rev{
\subsection{Real-World Experimental Details}
\label{app:real_world_details}

\noindent\textbf{Dataset and Task.}
The Tidy-up dataset contains 512 real-robot demonstrations collected
at 30\,Hz, with 461 trajectories used for training and 51 held out
for model selection and offline evaluation.
Each observation consists of three RGB views, the 7-DoF robot joint
state, and the binary gripper state.
Actions are represented by absolute joint targets and a binary
gripper command.
The demonstrations contain multi-stage sorting sequences in which
the robot places trash into a trash bin and other objects into
a storage basket.

The Tidy-up evaluation uses three trash items (an aluminum can,
crumpled paper, and a milk carton) and six non-trash objects
(a dinosaur toy, blue block, blue bowl, eggplant, pink cup,
and green pepper).
Each configuration contains two trash items and two non-trash objects. We select 40 distinct combinations for evaluation, with varied object positions and arrangements.

The dynamic pick-and-place task requires the robot to grasp a toy
and place it into a fixed basket.
Each trial follows a perturbation schedule with four successive
target-position displacements during grasping.
All target positions remain within the reachable workspace.
Each policy is evaluated over 20 trials under the same perturbation
procedure and task time limit.

\noindent\textbf{Training and Policy.}
All three evaluated variants use the same observation and action
representations and predict 16-step action chunks.
The original $\pi_{0.5}$ is evaluated with 10 and 1 flow-integration
steps using the same trained checkpoint.
For $\pi_{0.5}$-SOAR, we initialize from a converged task-specific
$\pi_{0.5}$ checkpoint, freeze the PaliGemma vision-language backbone,
and train the SOAR action module.

Training is performed for 30K optimization steps using AdamW with
an effective batch size of 128 on two 80\,GB GPUs.
Frequency-based conditioning is applied to the continuous
joint-action channels, while the binary gripper channel is
supervised separately.
Temporal regularization is applied to the joint trajectories,
and event-centered sampling increases the proportion of samples
containing gripper-state transitions.

We use RTC with a maximum prefix delay of eight actions.
Previously committed actions are provided as a temporal prefix
when generating the next action chunk.
SOAR generates the resulting chunk with a single network
evaluation ($\mathrm{NFE}=1$).

Inference latency is measured for a complete 16-step action chunk
on the same deployment hardware.
Action-generation latency measures the computation time of the
action module, whereas end-to-end inference latency includes
vision-language processing and action generation.
We report latency in milliseconds per chunk (ms/chunk), rather
than amortized latency per action.

\noindent\textbf{Tidy-up Evaluation Metrics.}
An episode is considered successful only when all four objects
are correctly sorted into their designated receptacles.
The full-task success rate is computed as
\begin{equation}
    \mathrm{SR}_{\mathrm{tidy}}
    =
    \frac{1}{40}
    \sum_{i=1}^{40} y_i
    \times 100\%,
\end{equation}
where $y_i\in\{0,1\}$ denotes the full-task outcome for
configuration $i$.
All policies are evaluated on the same 40 configurations.
This criterion measures completion of the entire sorting
sequence rather than individual grasp or placement attempts.

\noindent\textbf{Dynamic Evaluation Metrics.}
A dynamic episode is considered successful only when the robot
grasps the target, transports it to the basket, releases it,
and the target remains stably inside the basket within the
task time limit.
Timeouts, unsuccessful placements, and safety terminations
are counted as failures.
The placement success rate is
\begin{equation}
    \mathrm{SR}_{\mathrm{dyn}}
    =
    \frac{1}{20}
    \sum_{i=1}^{20} z_i
    \times 100\%,
\end{equation}
where $z_i\in\{0,1\}$ indicates placement success in trial $i$.

An empty grasp is recorded when the gripper completes a
closing attempt without securing the target object.
Continuous closure is counted as one attempt, while dropping
an object after a successful grasp is recorded separately.

Let $n_i$ denote the number of empty grasps in trial $i$.
We report the total empty-grasp count, mean empty grasps
per trial, and sample standard deviation:
\begin{equation}
\begin{aligned}
    N_{\mathrm{empty}}
    &= \sum_{i=1}^{20} n_i, \\
    \bar{n}_{\mathrm{empty}}
    &= \frac{1}{20}\sum_{i=1}^{20} n_i, \\
    s_{\mathrm{empty}}
    &= \sqrt{
        \frac{1}{19}
        \sum_{i=1}^{20}
        (n_i-\bar{n}_{\mathrm{empty}})^2
    }.
\end{aligned}
\end{equation}
All statistics include both successful and failed trials.
Empty-grasp counts are not normalized by the number of
target displacements.

\noindent\textbf{Detailed Experimental Results.}
Tables~\ref{tab:tidy_per_configuration} and
\ref{tab:dynamic_per_trial} report the individual outcomes
underlying Fig.~\ref{fig:real_world_results}.

On Tidy-up, the 10-step $\pi_{0.5}$, 1-step $\pi_{0.5}$,
and $\pi_{0.5}$-SOAR complete 30, 28, and 33 of the
40 configurations, respectively.
Compared with the 10-step baseline, SOAR succeeds on six
configurations where the baseline fails, while the reverse
occurs on three configurations.
Against the 1-step baseline, the corresponding counts
are seven and two.

On dynamic pick-and-place, the three policies succeed in
16, 13, and 17 of the 20 trials, respectively.
Their total empty-grasp counts are 19, 24, and 7, yielding
mean $\pm$ sample standard deviation of
$0.95\pm0.83$, $1.20\pm1.01$, and $0.35\pm0.49$
empty grasps per trial.
These per-configuration and per-trial results provide
a detailed account of the observed differences in
task success and unsuccessful grasp attempts.


\begin{table}[tbp]
\centering
\caption{
Per-configuration outcomes on the real-world Tidy-up task.
Each configuration contains two trash items and two non-trash objects,
separated by ``/''.
The columns 10, 1, and SOAR denote
$\pi_{0.5}$ (NFE=10), $\pi_{0.5}$ (NFE=1),
and $\pi_{0.5}$-SOAR (NFE=1), respectively.
1 indicates full-task success and 0 indicates failure.
The overall success rates are 75.0\%, 70.0\%, and 82.5\%.
}
\label{tab:tidy_per_configuration}

\scriptsize
\setlength{\tabcolsep}{2pt}
\renewcommand{\arraystretch}{0.95}

\resizebox{\textwidth}{!}{%
\begin{tabular}{@{}c l c c c @{\hspace{9pt}} c l c c c@{}}
\toprule
\textbf{ID} & \textbf{Trash / Non-trash Objects}
& \textbf{10} & \textbf{1} & \textbf{SOAR}
& \textbf{ID} & \textbf{Trash / Non-trash Objects}
& \textbf{10} & \textbf{1} & \textbf{SOAR} \\
\midrule

01 & Aluminum can, crumpled paper / Dinosaur toy, pink cup
& 0 & 0 & 1 &
21 & Crumpled paper, milk carton / Blue bowl, pink cup
& 1 & 1 & 1 \\

02 & Aluminum can, milk carton / Dinosaur toy, green pepper
& 1 & 0 & 1 &
22 & Aluminum can, crumpled paper / Blue block, pink cup
& 1 & 1 & 1 \\

03 & Crumpled paper, milk carton / Dinosaur toy, blue block
& 1 & 1 & 1 &
23 & Aluminum can, milk carton / Blue block, eggplant
& 0 & 0 & 0 \\

04 & Aluminum can, crumpled paper / Blue bowl, pink cup
& 0 & 0 & 1 &
24 & Crumpled paper, milk carton / Blue block, green pepper
& 1 & 1 & 1 \\

05 & Aluminum can, milk carton / Blue block, green pepper
& 1 & 0 & 0 &
25 & Aluminum can, crumpled paper / Blue block, eggplant
& 1 & 1 & 1 \\

06 & Crumpled paper, milk carton / Blue bowl, green pepper
& 1 & 1 & 1 &
26 & Aluminum can, milk carton / Dinosaur toy, blue bowl
& 0 & 0 & 1 \\

07 & Aluminum can, crumpled paper / Blue block, blue bowl
& 1 & 1 & 1 &
27 & Crumpled paper, milk carton / Blue block, pink cup
& 1 & 1 & 1 \\

08 & Aluminum can, milk carton / Blue bowl, eggplant
& 0 & 0 & 0 &
28 & Aluminum can, crumpled paper / Dinosaur toy, blue bowl
& 0 & 1 & 0 \\

09 & Crumpled paper, milk carton / Dinosaur toy, green pepper
& 1 & 1 & 1 &
29 & Aluminum can, milk carton / Blue bowl, pink cup
& 1 & 1 & 1 \\

10 & Aluminum can, crumpled paper / Eggplant, green pepper
& 1 & 1 & 1 &
30 & Crumpled paper, milk carton / Dinosaur toy, pink cup
& 1 & 1 & 1 \\

11 & Aluminum can, milk carton / Eggplant, green pepper
& 1 & 0 & 1 &
31 & Aluminum can, crumpled paper / Pink cup, green pepper
& 1 & 1 & 1 \\

12 & Crumpled paper, milk carton / Pink cup, green pepper
& 1 & 1 & 1 &
32 & Aluminum can, milk carton / Eggplant, pink cup
& 1 & 1 & 1 \\

13 & Aluminum can, crumpled paper / Dinosaur toy, green pepper
& 1 & 1 & 1 &
33 & Crumpled paper, milk carton / Eggplant, green pepper
& 0 & 0 & 0 \\

14 & Aluminum can, milk carton / Blue block, pink cup
& 1 & 1 & 1 &
34 & Aluminum can, crumpled paper / Blue block, green pepper
& 1 & 1 & 1 \\

15 & Crumpled paper, milk carton / Blue bowl, eggplant
& 0 & 1 & 1 &
35 & Aluminum can, milk carton / Pink cup, green pepper
& 1 & 1 & 1 \\

16 & Aluminum can, crumpled paper / Blue bowl, green pepper
& 1 & 1 & 1 &
36 & Crumpled paper, milk carton / Blue block, eggplant
& 1 & 0 & 1 \\

17 & Aluminum can, milk carton / Dinosaur toy, blue block
& 1 & 1 & 1 &
37 & Aluminum can, crumpled paper / Dinosaur toy, eggplant
& 1 & 1 & 0 \\

18 & Crumpled paper, milk carton / Dinosaur toy, blue bowl
& 1 & 0 & 0 &
38 & Aluminum can, milk carton / Blue block, blue bowl
& 0 & 1 & 1 \\

19 & Aluminum can, crumpled paper / Blue bowl, eggplant
& 0 & 1 & 1 &
39 & Crumpled paper, milk carton / Eggplant, pink cup
& 1 & 1 & 1 \\

20 & Aluminum can, milk carton / Dinosaur toy, pink cup
& 1 & 1 & 1 &
40 & Crumpled paper, milk carton / Dinosaur toy, eggplant
& 1 & 0 & 1 \\

\bottomrule
\end{tabular}%
}
\end{table}


\begin{table}[tbp]
\centering
\caption{
Per-trial results on dynamic pick-and-place.
EG denotes the number of empty grasps, and S indicates
placement success (1) or failure (0).
Each policy is evaluated over 20 trials with four
target displacements during grasping.
}
\label{tab:dynamic_per_trial}

\scriptsize
\setlength{\tabcolsep}{3pt}
\renewcommand{\arraystretch}{1.05}

\resizebox{\textwidth}{!}{%
\begin{tabular}{@{}l c *{20}{c} c@{}}
\toprule

\textbf{Policy}
& \textbf{Metric}
& \multicolumn{20}{c}{\textbf{Trial Index}}
& \textbf{Total} \\

\cmidrule(lr){3-22}

& & 01 & 02 & 03 & 04 & 05
  & 06 & 07 & 08 & 09 & 10
  & 11 & 12 & 13 & 14 & 15
  & 16 & 17 & 18 & 19 & 20 & \\

\midrule

$\pi_{0.5}$ (NFE=10)
& EG
& 2 & 1 & 1 & 1 & 0
& 0 & 1 & 2 & 1 & 1
& 1 & 0 & 0 & 1 & 1
& 1 & 0 & 2 & 3 & 0
& \textbf{19} \\

& S
& 0 & 1 & 1 & 1 & 1
& 1 & 1 & 0 & 1 & 1
& 1 & 1 & 1 & 1 & 1
& 1 & 1 & 0 & 0 & 1
& \textbf{16} \\

\midrule

$\pi_{0.5}$ (NFE=1)
& EG
& 2 & 0 & 1 & 1 & 2
& 0 & 1 & 3 & 1 & 0
& 1 & 0 & 3 & 2 & 2
& 2 & 1 & 0 & 2 & 0
& \textbf{24} \\

& S
& 0 & 1 & 1 & 1 & 0
& 1 & 1 & 0 & 1 & 1
& 1 & 1 & 0 & 0 & 0
& 0 & 1 & 1 & 1 & 1
& \textbf{13} \\

\midrule

$\pi_{0.5}$-SOAR (NFE=1)
& EG
& 0 & 0 & 1 & 0 & 1
& 1 & 0 & 1 & 0 & 0
& 0 & 0 & 0 & 1 & 1
& 0 & 0 & 0 & 1 & 0
& \textbf{7} \\

& S
& 1 & 1 & 0 & 1 & 1
& 0 & 1 & 1 & 1 & 1
& 1 & 1 & 1 & 1 & 0
& 1 & 1 & 1 & 1 & 1
& \textbf{17} \\

\bottomrule
\end{tabular}%
}

\vspace{3pt}

{\footnotesize
Mean EG $\pm$ SD:
$0.95\pm0.83$ ($\pi_{0.5}$, NFE=10),
$1.20\pm1.01$ ($\pi_{0.5}$, NFE=1),
and $0.35\pm0.49$ ($\pi_{0.5}$-SOAR).
}

\end{table}
}

\rev{

\subsection{Other Experimental Details}
\label{app:other_experimental_details}

\subsubsection{Temporal Structure Analysis}
\label{app:temporal_structure_eval}

\noindent\textbf{Evaluation Protocol.}
We evaluate temporal trajectory discrepancies using
saved final-checkpoint evaluation records without
additional policy rollouts. Robot motions are
reconstructed from recorded actions and compared
with expert reference motions from the corresponding
scenarios. For Adroit, we use palm XYZ positions;
for Meta-World, we use end-effector XYZ positions.
All metrics are computed over complete trajectories.

For DP3, we compare NFE=1 and NFE=10 across
53 tasks, including three Adroit tasks and
50 Meta-World tasks. Each inference setting
contains three training seeds and one recorded
closed-loop episode per task and seed, yielding
318 trajectories in total.

For SOAR, we examine the motions before and after
action calibration using saved action records.
Pre-calibration motions are reconstructed from
the recorded base actions rather than obtained
through separate closed-loop policy execution.

\noindent\textbf{Metrics.}
Let $P=\{p_t\}_{t=1}^{L}$ and
$P^\star=\{p_t^\star\}_{t=1}^{L}$
denote temporally aligned robot and expert
reference trajectories, where
$p_t,p_t^\star\in\mathbb{R}^{3}$.
We subtract the temporal mean of each trajectory:
$\widetilde{P}=P-\bar{P}$, $\widetilde{P}^{\star}=P^{\star}-\bar{P}^{\star}$.

Temporal shape error (TSE) measures the position
MSE between the centered trajectories:
$\mathrm{TSE}=\operatorname{MSE}(\widetilde{P},\widetilde{P}^{\star})$.

Amplitude-spectrum error (ASE) measures the
difference between their Fourier magnitude spectra:
$\mathrm{ASE}=\operatorname{MSE}(|\mathcal{F}(\widetilde{P})|,|\mathcal{F}(\widetilde{P}^{\star})|)$,
where $\mathcal{F}$ denotes the Fourier transform.
ASE ignores spectral phase, whereas TSE retains
temporal correspondence. Lower values indicate
smaller discrepancies from expert reference motion.

For each metric, task-group statistics are reported
as mean $\pm$ sample standard deviation over
three training seeds. The overall mean is
recomputed from the reported task-group means,
weighted by the number of tasks:
\begin{equation}
\begin{aligned}
    \overline{E}_{\mathrm{all}}
    =\frac{1}{53}\big(
    &3\overline{E}_{\mathrm{Adroit}}
    +28\overline{E}_{\mathrm{Easy}}\\
    &+11\overline{E}_{\mathrm{Medium}}
    +6\overline{E}_{\mathrm{Hard}}
    +5\overline{E}_{\mathrm{VeryHard}}
    \big).
\end{aligned}
\end{equation}
The overall values are approximate because they
are calculated from rounded task-group means.
Overall standard deviations are omitted because
the seed-wise task-group means are required
to recompute them.

\noindent\textbf{DP3: One-step vs. Multi-step Generation.}
Table~\ref{tab:dp3_temporal_structure}
compares the trajectory errors of DP3 with
NFE=1 and NFE=10.
Increasing NFE from 1 to 10 reduces the overall
TSE from 2.206 to 2.057 and ASE from 0.996 to
0.852, corresponding to reductions of 6.8\%
and 14.4\%, respectively.
Both errors are lower at NFE=10 across all
five task groups.

These measurements indicate that fewer inference
steps are associated with larger trajectory-shape
and spectral discrepancies in this DP3 evaluation.

\begin{table}[tbp]
    \centering
    \caption{
        Temporal trajectory errors of DP3 under
        NFE=1 and NFE=10 across 53 tasks.
        TSE and ASE are in units of
        $10^{-3}\,\mathrm{m}^{2}$.
        Task-group results are mean $\pm$
        sample standard deviation over three
        training seeds. Overall reports the
        task-weighted mean recomputed from the
        displayed group means. Lower is better.
    }
    \label{tab:dp3_temporal_structure}
    \vspace{1mm}

    \setlength{\tabcolsep}{12pt}
    \renewcommand{\arraystretch}{1.1}

    \resizebox{\textwidth}{!}{%
    \begin{tabular}{lcccc}
        \toprule
        & \multicolumn{2}{c}{\textbf{TSE}}
        & \multicolumn{2}{c}{\textbf{ASE}} \\
        \cmidrule(lr){2-3}
        \cmidrule(lr){4-5}

        \textbf{Task Group}
        & \textbf{NFE=1}
        & \textbf{NFE=10}
        & \textbf{NFE=1}
        & \textbf{NFE=10} \\
        \midrule

        Adroit (3)
        & $0.792 \pm 0.060$
        & $0.757 \pm 0.048$
        & $0.281 \pm 0.045$
        & $0.248 \pm 0.015$ \\

        Easy (28)
        & $2.373 \pm 0.286$
        & $2.273 \pm 0.184$
        & $0.984 \pm 0.058$
        & $0.845 \pm 0.037$ \\

        Medium (11)
        & $2.225 \pm 0.638$
        & $1.918 \pm 0.503$
        & $1.168 \pm 0.305$
        & $0.998 \pm 0.352$ \\

        Hard (6)
        & $0.404 \pm 0.024$
        & $0.389 \pm 0.052$
        & $0.260 \pm 0.016$
        & $0.259 \pm 0.031$ \\

        Very Hard (5)
        & $4.236 \pm 1.734$
        & $3.930 \pm 2.012$
        & $1.992 \pm 1.395$
        & $1.645 \pm 1.668$ \\

        \midrule

        \textbf{Overall (53)}
        & $\mathbf{2.206}$
        & $\mathbf{2.057}$
        & $\mathbf{0.996}$
        & $\mathbf{0.852}$ \\

        \bottomrule
    \end{tabular}%
    }
\end{table}

\noindent\textbf{SOAR: Effect of Action Calibration.}
Table~\ref{tab:soar_temporal_structure}
compares the reconstructed trajectory errors
before and after action calibration.
Calibration reduces the overall TSE from
3.438 to 2.092, corresponding to a reduction
of 39.2\%. ASE decreases from 1.483 to 0.936,
a reduction of 36.9\%.

Both TSE and ASE decrease after calibration
in all five task groups. These results show
that the calibrated actions have smaller
trajectory-shape and spectral discrepancies
relative to expert reference motion under
the recorded evaluation conditions.

\begin{table}[tbp]
    \centering
    \caption{
        Temporal trajectory errors of SOAR before
        and after action calibration.
        TSE and ASE are in units of
        $10^{-3}\,\mathrm{m}^{2}$.
        Task-group results are mean $\pm$
        sample standard deviation over three
        training seeds. Overall reports the
        task-weighted mean recomputed from the
        displayed group means. Lower is better.
    }
    \label{tab:soar_temporal_structure}
    \vspace{1mm}

    \setlength{\tabcolsep}{12pt}
    \renewcommand{\arraystretch}{1.1}

    \resizebox{\textwidth}{!}{%
    \begin{tabular}{lcccc}
        \toprule
        & \multicolumn{2}{c}{\textbf{TSE}}
        & \multicolumn{2}{c}{\textbf{ASE}} \\
        \cmidrule(lr){2-3}
        \cmidrule(lr){4-5}

        \textbf{Task Group}
        & \textbf{Before AC}
        & \textbf{After AC}
        & \textbf{Before AC}
        & \textbf{After AC} \\
        \midrule

        Adroit (3)
        & $4.118 \pm 3.235$
        & $1.665 \pm 0.881$
        & $2.104 \pm 1.618$
        & $0.525 \pm 0.265$ \\

        Easy (28)
        & $2.579 \pm 0.433$
        & $1.720 \pm 0.136$
        & $1.153 \pm 0.091$
        & $0.760 \pm 0.014$ \\

        Medium (11)
        & $3.915 \pm 0.458$
        & $1.563 \pm 0.229$
        & $1.774 \pm 0.412$
        & $1.116 \pm 0.177$ \\

        Hard (6)
        & $1.001 \pm 0.058$
        & $0.878 \pm 0.444$
        & $0.621 \pm 0.027$
        & $0.497 \pm 0.239$ \\

        Very Hard (5)
        & $9.714 \pm 1.216$
        & $7.047 \pm 1.545$
        & $3.355 \pm 0.628$
        & $2.298 \pm 0.213$ \\

        \midrule

        \textbf{Overall (53)}
        & $\mathbf{3.438}$
        & $\mathbf{2.092}$
        & $\mathbf{1.483}$
        & $\mathbf{0.936}$ \\

        \bottomrule
    \end{tabular}%
    }
\end{table}

\subsubsection{Independent Final-Checkpoint Evaluation}
\label{app:final_checkpoint_eval}

\noindent\textbf{Evaluation Protocol.}
The main 3D benchmark results follow the periodic-evaluation
protocol used in prior work, including MP1. Models are evaluated
every 200 training epochs, and the five highest evaluation scores
are averaged for each training seed. These reported scores come
from the same evaluations used for checkpoint selection; no
additional held-out rollouts were used for the original Top-5
results. The simulation environment was reset for each rollout,
but initial states were not explicitly matched across checkpoints.

To assess the effect of checkpoint selection, we additionally
evaluate the final training checkpoints of SOAR and MP1 on
separate evaluation episodes that were not used for checkpoint
selection. No checkpoint is selected using its periodic evaluation
score. The two policies are evaluated with matched initial
observation configurations. We use three independent training
seeds and report the mean success rate and sample standard
deviation across seeds. The evaluation covers all 53 tasks:
three Adroit tasks and 50 Meta-World tasks.

\noindent\textbf{Results.}
Table~\ref{tab:independent_final_checkpoint} reports the
independent final-checkpoint results. On Adroit, SOAR obtains
$77.11\pm3.67\%$, compared with $71.11\pm3.85\%$ for MP1.
On Meta-World, SOAR achieves higher mean success rates on
Easy ($94.05\pm2.06\%$ vs. $91.90\pm0.82\%$),
Medium ($70.91\pm4.81\%$ vs. $63.64\pm1.82\%$),
Hard ($77.78\pm9.62\%$ vs. $66.67\pm6.94\%$), and
Very Hard ($72.00\pm5.25\%$ vs. $64.00\pm9.62\%$).
SOAR has a higher mean in all five groups under this
final-checkpoint evaluation. Thus, the advantage over MP1
is also observed without selecting checkpoints by their
periodic evaluation scores.

\begin{table}[tbp]
    \centering
    \caption{
        Independent evaluation of final training checkpoints
        on 53 Adroit and Meta-World tasks. Evaluation episodes
        were not used for periodic checkpoint selection, and
        initial observation configurations were matched between
        SOAR and MP1. Success rates (\%) are mean $\pm$ sample
        standard deviation across three training seeds.
    }
    \label{tab:independent_final_checkpoint}
    \vspace{1mm}
    \setlength{\tabcolsep}{5pt}
    \renewcommand{\arraystretch}{1.10}

    \resizebox{\columnwidth}{!}{%
    \begin{tabular}{lccccc}
        \toprule
        \textbf{Method}
        & \textbf{Adroit (3)}
        & \textbf{Easy (28)}
        & \textbf{Medium (11)}
        & \textbf{Hard (6)}
        & \textbf{Very Hard (5)} \\
        \midrule

        MP1
        & $71.11\pm3.85$
        & $91.90\pm0.82$
        & $63.64\pm1.82$
        & $66.67\pm6.94$
        & $64.00\pm9.62$ \\

        \textbf{SOAR}
        & $\mathbf{77.11\pm3.67}$
        & $\mathbf{94.05\pm2.06}$
        & $\mathbf{70.91\pm4.81}$
        & $\mathbf{77.78\pm9.62}$
        & $\mathbf{72.00\pm5.25}$ \\

        \bottomrule
    \end{tabular}%
    }
\end{table}

\subsubsection{Structure-Only Prediction and Two-Branch Evaluation}
\label{app:two_branch_eval}

\noindent\textbf{Evaluation Protocol.}
We evaluate three action outputs from the final training
checkpoints of SOAR: (i) the inverse-DCT reconstruction
of the structure head prediction, (ii) the uncalibrated
base action produced by the one-step generator, and
(iii) the final action after spectral calibration.
All three outputs are evaluated under the same
diagnostic protocol within each benchmark.
This comparison examines the respective roles of
structure prediction, action generation, and calibration.

\noindent\textbf{Structure Prediction and Calibration.}
The observation-conditioned structure head predicts
$\mathbf{C}^{s}=H_{\psi}(\mathbf{c})$, supervised by
the DCT coefficients of expert actions.
In the 3D experiments, the complete set of DCT
coefficients is used ($K=T=4$), allowing the inverse
discrete cosine transform (IDCT) to reconstruct
an action sequence
$\mathbf{A}_{\mathrm{struct}}=
\operatorname{IDCT}(\mathbf{C}^{s})$.
However, the structure head is not designed to
independently generate reliable executable trajectories.
Its primary role is to learn the temporal structure
of expert actions and provide explicit guidance
for subsequent action generation.

Specifically, the predicted frequency bands are
embedded into the observation condition to guide
the one-step generator, which produces the base
action $\mathbf{A}_{\mathrm{base}}$.
The same structure prediction is further used to
calibrate the generated action.
For the complete orthonormal DCT, the calibration
in Eq.~\eqref{eq:spectral_action_calibration}
can be written exactly as
\begin{equation}
    \widehat{\mathbf{A}}
    = (1-\alpha)\mathbf{A}_{\mathrm{base}}
    + \alpha\mathbf{A}_{\mathrm{struct}},
    \qquad \alpha=\tanh(\rho).
    \label{eq:two_branch_scalar_fusion}
\end{equation}
This formulation shows that calibration corresponds
to a scalar blend in action space.
The structure head and generator nevertheless serve
different purposes: the former provides temporal
guidance, while the latter produces the base
executable action.

\noindent\textbf{Results.}
Table~\ref{tab:two_branch_eval} compares the
direct execution performance of the three outputs.
The structure-only reconstruction obtained by
applying IDCT to the predicted coefficients
achieves success rates of only 17.67\% on Adroit
and 24.03\% on Meta-World.
Although the full DCT coefficients can reconstruct
a complete action sequence, these low success rates
indicate that the structure prediction alone
does not constitute a reliable executable policy.
Instead, its primary purpose is to capture
task-relevant temporal structure that supports
action generation.

In contrast, the one-step generator produces
substantially more reliable base actions,
achieving 59.22\% and 69.07\% on Adroit
and Meta-World, respectively.
This demonstrates the importance of the generator
in converting structure-conditioned observations
into executable action trajectories.
Applying spectral calibration further improves
success rates to 74.56\% and 80.60\%,
corresponding to gains of 15.34 and 11.53
percentage points over the uncalibrated
base actions.

These results highlight the complementary roles
of the three components: the structure head
provides temporal guidance, the one-step generator
produces executable actions, and spectral
calibration refines the generated trajectories
using the predicted structure.

\subsubsection{Deterministic Regression and Sample Diversity Analysis}
\label{app:sample_diversity}
\label{app:deterministic_regression}

\noindent\textbf{Deterministic Regression Baseline.}
We compare SOAR with an MSE-trained deterministic policy
using the same observation encoder and action U-Net.
The baseline conditions on the two most recent observations
and predicts a four-step action sequence from fixed
zero-action and time inputs.
Unlike SOAR, it does not use stochastic noise,
structure prediction, multi-band conditioning,
action calibration, or MeanFlow training.
Both policies require one network evaluation
($\mathrm{NFE}=1$).

We evaluate 20 tasks (3 Adroit, 11 Meta-World Medium,
and 6 Hard) with three training seeds.
Adroit and Meta-World are trained for 3,000 and
1,000 epochs, respectively, with evaluation every
200 epochs.
Each evaluation uses 50 episodes on Adroit and
20 on Meta-World.
We average the five highest evaluation scores per
task and seed, then report the mean and sample
standard deviation across seeds.
On Meta-World, all five evaluations are included.

\noindent\textbf{Regression Results.}
Table~\ref{tab:deterministic_regression} reports
the comparison.
Deterministic regression achieves
$63.52\pm1.25\%$ average success across 20 tasks,
compared with $71.9\%$ for SOAR.
The improvement is 8.38 percentage points overall,
including 9.91 and 9.28 points on Meta-World Medium
and Hard, respectively.
The direct-action generative baseline in
Table~\ref{tab:ablation_adroit} achieves $69.3\%$.
These comparisons distinguish deterministic regression
from one-step generative prediction.
The gain over regression reflects the complete SOAR
design rather than any individual component.

\noindent\textbf{Conditional Sampling Protocol.}
We further compare deterministic regression and SOAR
on Meta-World \texttt{basketball}.
Using SOAR trained with seed 0, we select five fixed
observations and generate 32 samples per observation
from different Gaussian noise initializations.
The base and calibrated actions use identical noise
initializations.
We evaluate the three action steps retained for
execution without additional environment rollouts.
Deterministic regression produces identical predictions
for repeated inputs.
Results are reported as mean $\pm$ standard deviation
across the five observations.

\noindent\textbf{Diversity Metrics.}
Let $\{\mathbf{A}_i\}_{i=1}^{M}$ denote $M=32$
samples for a fixed observation, with
$\mathbf{A}_i\in\mathbb{R}^{T_e\times D_a}$,
$T_e=3$, and action dimension $D_a$.
Overall and temporal-shape diversity are defined as
\begin{equation}
\begin{aligned}
D_{\mathrm{all}}
&=\frac{2}{M(M-1)}
\sum_{i<j}
\frac{\|\mathbf{A}_i-\mathbf{A}_j\|_F^2}{T_eD_a},
\\
D_{\mathrm{shape}}
&=\frac{2}{M(M-1)}
\sum_{i<j}
\frac{\|\widetilde{\mathbf{A}}_i-
\widetilde{\mathbf{A}}_j\|_F^2}{T_eD_a},
\end{aligned}
\end{equation}
where
$\widetilde{\mathbf{A}}_i=
\mathbf{A}_i-\operatorname{Mean}_t(\mathbf{A}_i)$.
We additionally measure pairwise variation in
consecutive-action differences.

Prediction errors relative to the recorded
expert action $\mathbf{A}^{\star}$ are
\begin{equation}
\begin{aligned}
E_{\mathrm{mean}}
&=\frac{1}{M}\sum_{i=1}^{M}
\frac{\|\mathbf{A}_i-\mathbf{A}^{\star}\|_F^2}
{T_eD_a},
\\
E_{\mathrm{best}}
&=\min_{i=1,\ldots,M}
\frac{\|\mathbf{A}_i-\mathbf{A}^{\star}\|_F^2}
{T_eD_a}.
\end{aligned}
\end{equation}
For deterministic regression, all pairwise distances
are zero and $E_{\mathrm{best}}=E_{\mathrm{mean}}$.

\noindent\textbf{Diversity Results.}
Table~\ref{tab:sample_diversity} compares
deterministic regression with SOAR before and after
action calibration.
On \texttt{basketball}, calibration retains
90.5\% of overall pairwise diversity and
97.6\% of temporal-shape diversity.
Mean expert-reference MSE decreases from
46.197 to 0.733, a reduction of 98.4\%.
Calibrated SOAR also achieves lower reference
MSE than deterministic regression (0.733 versus 22.880).

These measurements show that SOAR retains nonzero
conditional sample variation after calibration,
unlike deterministic regression.


\begin{table}[tbp]
    \centering
    \begin{minipage}[t]{0.48\textwidth}
        \vspace{0pt}
        \centering
        \caption{Direct execution of three SOAR outputs using final training checkpoints. IDCT denotes the inverse discrete cosine transform. IDCT (Structure) reconstructs actions directly from the predicted structure coefficients without the generator. Base Action and Calibrated denote outputs before and after spectral calibration. Results are success rates (\%).
        }
        \label{tab:two_branch_eval}
        \vspace{1mm}
        \begingroup
        \fontsize{8pt}{10pt}\selectfont
        \setlength{\tabcolsep}{4pt}
        \renewcommand{\arraystretch}{1.15}
        \begin{tabular*}{\linewidth}
            {@{\extracolsep{\fill}}lcc@{}}
            \toprule
            \textbf{Action Output}
            & \textbf{Adroit}
            & \textbf{Meta-World} \\
            \midrule
            IDCT (Structure)
            & $17.67$
            & $24.03$ \\
            Base Action
            & $59.22$
            & $69.07$ \\
            Calibrated
            & $\mathbf{74.56}$
            & $\mathbf{80.60}$ \\
            \bottomrule
        \end{tabular*}
        \endgroup
    \end{minipage}%
    \hfill%
    \begin{minipage}[t]{0.48\textwidth}
        \vspace{0pt}
        \centering
        \caption{Deterministic regression versus SOAR on 20 manipulation tasks. Results are success rates (\%). Regression reports mean $\pm$ standard deviation across three seeds.
        }
        \label{tab:deterministic_regression}
        \vspace{1mm}
        \begingroup
        \fontsize{8pt}{10pt}\selectfont
        \setlength{\tabcolsep}{4pt}
        \renewcommand{\arraystretch}{1.15}
        \begin{tabular*}{\linewidth}
            {@{\extracolsep{\fill}}lcc@{}}
            \toprule
            \textbf{Task Group}
            & \textbf{Regression}
            & \textbf{SOAR} \\
            \midrule
            Adroit (3)
            & $78.91 \pm 2.70$
            & $\mathbf{79.8}$ \\
            MW Medium (11)
            & $64.39 \pm 1.65$
            & $\mathbf{74.3}$ \\
            MW Hard (6)
            & $54.22 \pm 2.28$
            & $\mathbf{63.5}$ \\
            \midrule
            Overall (20)
            & $63.52 \pm 1.25$
            & $\mathbf{71.9}$ \\
            \bottomrule
        \end{tabular*}
        \endgroup
    \end{minipage}%
\end{table}

\begin{table}[tbp]
    \centering
    \caption{
        Conditional sample diversity on
        Meta-World \texttt{basketball}.
        Results are mean $\pm$ standard deviation
        over five observations, scaled by $10^{-3}$.
        AC denotes action calibration.
    }
    \label{tab:sample_diversity}
    \vspace{1mm}

    \begingroup
    \fontsize{8pt}{10pt}\selectfont
    \setlength{\tabcolsep}{3pt}
    \renewcommand{\arraystretch}{1.10}

    \resizebox{\columnwidth}{!}{%
    \begin{tabular}{lccc}
        \toprule
        \textbf{Metric}
        & \textbf{Regression}
        & \textbf{Before AC}
        & \textbf{After AC} \\
        \midrule

        Overall pairwise distance
        & $0 \pm 0$
        & $4.238 \pm 3.724$
        & $3.834 \pm 2.827$ \\

        Temporal-shape distance
        & $0 \pm 0$
        & $1.993 \pm 1.394$
        & $1.945 \pm 1.276$ \\

        Action-difference diversity
        & $0 \pm 0$
        & $3.061 \pm 2.128$
        & $2.991 \pm 1.925$ \\

        Mean reference MSE
        & $22.880 \pm 47.746$
        & $46.197 \pm 81.401$
        & $\mathbf{0.733 \pm 0.688}$ \\

        Best-of-32 MSE
        & $22.880 \pm 47.746$
        & $45.196 \pm 80.067$
        & $\mathbf{0.570 \pm 0.600}$ \\

        \bottomrule
    \end{tabular}%
    }
    \endgroup
\end{table}

Thus, calibration adjusts the differences between
samples without replacing stochastic action generation.
The measured results show that this correction
substantially reduces reference-action error while
retaining most of the sampled action variation.

\subsubsection{Frequency Band Partition Ablation}
\label{app:frequency_band_ablation}

\noindent\textbf{Evaluation Setup.}
We evaluate the effect of frequency-band partitioning
on three Adroit tasks and 17 Meta-World tasks,
including 11 Medium and six Hard tasks.
For these 3D manipulation benchmarks, the action
horizon is $T=4$, and the structure head predicts
all four DCT coefficients ($K=4$).
We compare three frequency partitions:
$\text{2-band:}\ \mathcal{B}_1=\{0,1\},\ \mathcal{B}_2=\{2,3\}$; $\text{3-band:}\ \mathcal{B}_1=\{0,1\},\ \mathcal{B}_2=\{2\},\ \mathcal{B}_3=\{3\}$; $\text{4-band:}\ \mathcal{B}_1=\{0\},\ \mathcal{B}_2=\{1\},\ \mathcal{B}_3=\{2\},\ \mathcal{B}_4=\{3\}$.

All configurations retain the same DCT coefficients.
The frequency groups and their corresponding
conditioning projections and gates are adjusted,
while the structure prediction and action calibration
formulations remain unchanged.
The original SOAR uses the three-band partition.

We follow the component ablation protocol.
Each task is trained independently
using 10 expert demonstrations and three random seeds.
For each task and seed, we average the five highest
periodic evaluation success rates.
Task-group results are reported as mean $\pm$
sample standard deviation across seeds.
All variants retain one-step inference
($\mathrm{NFE}=1$).

\begin{table}[tbp]
    \centering
    \caption{
        Effect of frequency-band partitioning
        on Adroit and Meta-World.
        All configurations use $T=K=4$ and retain
        the same DCT coefficients.
        Success rates (\%) are reported as mean
        $\pm$ sample standard deviation over three
        training seeds.
        Overall is the task-weighted mean across
        20 tasks, computed from the reported group
        means. The three-band configuration corresponds
        to the original SOAR.
        Bold denotes the highest mean in each column.
    }
    \label{tab:frequency_band_ablation}
    \vspace{1mm}

    \setlength{\tabcolsep}{12pt}
    \renewcommand{\arraystretch}{1.1}

    \resizebox{\textwidth}{!}{%
    \begin{tabular}{lcccc}
        \toprule
        &
        \multicolumn{1}{c}{\textbf{Adroit (3)}}
        & \multicolumn{2}{c}{\textbf{Meta-World (17)}}
        & \\
        \cmidrule(lr){2-2}
        \cmidrule(lr){3-4}

        \textbf{Partition}
        & \textbf{Adroit}
        & \textbf{Medium (11)}
        & \textbf{Hard (6)}
        & \textbf{Overall (20)} \\
        \midrule

        2-band
        & $78.82 \pm 1.55$
        & $72.82 \pm 1.99$
        & $63.28 \pm 2.30$
        & $70.9$ \\

        \textbf{3-band (SOAR)}
        & $\mathbf{79.8 \pm 2.9}$
        & $\mathbf{74.3 \pm 0.6}$
        & $\mathbf{63.5 \pm 1.8}$
        & $\mathbf{71.9}$ \\

        4-band
        & $78.80 \pm 1.05$
        & $73.45 \pm 0.64$
        & $63.28 \pm 2.67$
        & $71.2$ \\

        \bottomrule
    \end{tabular}%
    }
\end{table}

\noindent\textbf{Results.}
Table~\ref{tab:frequency_band_ablation}
compares the three frequency partitions.
The original three-band SOAR achieves
$79.8\pm2.9\%$ on Adroit,
$74.3\pm0.6\%$ on Meta-World Medium,
and $63.5\pm1.8\%$ on Meta-World Hard.
Its task-weighted average across 20 tasks is
$71.9\%$, compared with $70.9\%$ for
2-band and $71.2\%$ for 4-band.

The three-band configuration achieves the
highest mean success rate in all three
task groups. On Meta-World Medium, it improves
over 2-band and 4-band by 1.48 and 0.85
percentage points, respectively.
The differences on Hard are smaller,
with a gain of 0.22 points over both alternatives. These results support the three-band configuration
for the evaluated short-horizon tasks.

\subsubsection{MP1-Augmented Baseline Comparisons}
\label{app:mp1_augmented_baselines}

\noindent\textbf{Evaluation Setup.}
To assess whether SOAR's gains mainly arise from
the MP1/MeanFlow training components, we augment
DP, DP3, CP, and FlowPolicy with the MeanFlow
objective, guidance target, and Dispersive Loss
used in our local MP1 implementation.

Each augmented baseline retains its observation
encoder and action-generation backbone.
We replace the original diffusion or consistency
training objective with MeanFlow training.
Given an expert action sequence $\mathbf{a}$
and Gaussian noise
$\boldsymbol{\epsilon}\sim\mathcal{N}(0,I)$,
we construct
$\mathbf{z}_t=(1-t)\mathbf{a}+t\boldsymbol{\epsilon}$.

The network predicts the average velocity
$u_\theta(\mathbf{z}_t,t,r,\mathbf{c})$,
where $\mathbf{c}$ denotes the observation
features. The two time variables $t$ and $r$
are encoded independently, and their embeddings
are added before being combined with the
observation features. The MeanFlow target is
constructed using Jacobian-vector products (JVPs).

We further adopt the guidance target from
our local MP1 implementation:
$\mathbf{u}_{\mathrm{ref}}=\operatorname{sg}[u_\theta(\mathbf{z}_t,t,t,\mathbf{c})]$, $\widehat{\mathbf{v}}=2(\boldsymbol{\epsilon}-\mathbf{a})-\mathbf{u}_{\mathrm{ref}}$,
where $\operatorname{sg}$ denotes stop-gradient.
No condition dropout or additional unconditional
inference branch is used.

Dispersive Loss is applied to intermediate features
from the U-Net downsampling stages, excluding
the final stage. The training objective is
$\mathcal{L}=\mathcal{L}_{\mathrm{MeanFlow}}+0.5\mathcal{L}_{\mathrm{dispersive}}$.

At inference, the action sequence is recovered
from a single velocity prediction:
$\widehat{\mathbf{a}}=\boldsymbol{\epsilon}-u_\theta(\boldsymbol{\epsilon},1,0,\mathbf{c})$.
All augmented baselines therefore retain
one-step inference ($\mathrm{NFE}=1$).

DP + MP1 retains the point-cloud-adapted DP
PointNet encoder, robot-state encoder, and U-Net.
DP3 + MP1 retains the full DP3 encoder and U-Net,
while FlowPolicy + MP1 retains the official
FlowPolicy encoder and U-Net.

For CP, we adapt its student network to the same
point-cloud inputs and dual-time conditioning.
After removing the teacher network and distillation
objective, the resulting student-only implementation
is equivalent to our adapted DP3 + MP1 model.
We therefore report their shared result rather
than treating them as independent baselines.

We evaluate all methods on the same 37-task
benchmark, including three Adroit tasks and
34 Meta-World tasks (21 Easy, four Medium,
four Hard, and five Very Hard).
Results are reported as mean $\pm$ sample
standard deviation over three training seeds.

\begin{table}[tbp]
    \centering
    \caption{
        Comparison of SOAR with MP1-augmented
        baselines on 37 Adroit and Meta-World tasks.
        The augmented baselines retain their
        respective encoders and action-generation
        backbones while adopting MeanFlow/JVP training,
        the MP1 guidance target, and Dispersive Loss.
        CP denotes the adapted student-only model
        and shares the DP3 + MP1 result.
        All methods use $\mathrm{NFE}=1$.
        Success rates (\%) are reported as mean
        $\pm$ sample standard deviation over
        three training seeds.
        Bold indicates the highest mean in each column.
    }
    \label{tab:mp1_augmented_baselines}
    \vspace{1mm}

    \setlength{\tabcolsep}{8pt}
    \renewcommand{\arraystretch}{1.10}

    \resizebox{\textwidth}{!}{%
    \begin{tabular}{lcccccccc}
        \toprule
        & &
        \multicolumn{1}{c}{\textbf{Adroit (3)}}
        & \multicolumn{5}{c}{\textbf{Meta-World (34)}}
        & \\
        \cmidrule(lr){3-3}
        \cmidrule(lr){4-8}

        \textbf{Methods}
        & \textbf{NFE}
        & \textbf{Average}
        & \textbf{Easy (21)}
        & \textbf{Medium (4)}
        & \textbf{Hard (4)}
        & \textbf{Very Hard (5)}
        & \textbf{Average}
        & \textbf{Overall (37)} \\
        \midrule

        DP + MP1
        & 1
        & $60.8 \pm 9.8$
        & $91.0 \pm 3.5$
        & $70.0 \pm 18.9$
        & $\mathbf{67.5 \pm 18.5}$
        & $63.0 \pm 12.3$
        & $81.6 \pm 4.4$
        & $79.9 \pm 4.2$ \\

        DP3 + MP1 / CP
        & 1
        & $75.8 \pm 10.2$
        & $91.2 \pm 3.6$
        & $67.5 \pm 21.5$
        & $62.5 \pm 20.9$
        & $70.0 \pm 10.5$
        & $81.9 \pm 4.6$
        & $81.4 \pm 4.3$ \\

        FlowPolicy + MP1
        & 1
        & $\mathbf{86.7 \pm 7.3}$
        & $87.9 \pm 4.7$
        & $72.5 \pm 19.6$
        & $60.0 \pm 20.1$
        & $73.0 \pm 9.4$
        & $80.6 \pm 4.6$
        & $81.1 \pm 4.3$ \\

        \midrule

        \textbf{SOAR (Ours)}
        & 1
        & $79.8 \pm 2.9$
        & $\mathbf{92.3 \pm 0.5}$
        & $\mathbf{76.8 \pm 0.7}$
        & $63.5 \pm 2.2$
        & $\mathbf{76.9 \pm 1.8}$
        & $\mathbf{84.8 \pm 0.7}$
        & $\mathbf{84.4 \pm 0.5}$ \\

        \bottomrule
    \end{tabular}%
    }
\end{table}

\noindent\textbf{Results.}
Table~\ref{tab:mp1_augmented_baselines}
compares SOAR with the MP1-augmented baselines
under the same one-step inference setting.
DP + MP1, DP3 + MP1, and FlowPolicy + MP1
achieve overall success rates of
$79.9\pm4.2\%$, $81.4\pm4.3\%$, and
$81.1\pm4.3\%$, respectively.
SOAR achieves $84.4\pm0.5\%$, exceeding
these methods by 4.5, 3.0, and 3.3
percentage points.

On Meta-World, SOAR achieves $84.8\pm0.7\%$,
compared with $81.6\pm4.4\%$,
$81.9\pm4.6\%$, and $80.6\pm4.6\%$
for the three augmented baselines.
SOAR also obtains the highest mean success
rates on Easy, Medium, and Very Hard tasks.
The advantage is not uniform across all
task groups: FlowPolicy + MP1 performs
better on Adroit, while DP + MP1 performs
better on Meta-World Hard.

These results indicate that the additional gains arise from using the predicted temporal structure during action generation and calibration, rather than from the shared MP1 training components alone. These experiments support the effectiveness of SOAR's structure-guided one-step action generation beyond the adoption of MP1/MeanFlow training alone.

\subsubsection{Longer-Horizon VLA Ablations}
\label{app:long_horizon_ablation}

\noindent\textbf{Evaluation Setup.}
We further evaluate the contributions of SOAR's
components on LIBERO-LONG using a SOAR-integrated
VLA policy. The action prediction horizon is set
to $H=10$, with five actions executed per inference.
We compare the complete SOAR model with variants
without multi-band conditioning (MC), action
calibration (AC), or inference-endpoint alignment
(EA). Each variant removes the corresponding
component while retaining the remaining components.
All configurations use one-step action generation
($\mathrm{NFE}=1$).

\begin{table}[tbp]
    \centering
    \caption{
        Component ablations on LIBERO-LONG using
        a SOAR-integrated VLA policy.
        MC: multi-band conditioning;
        AC: action calibration;
        EA: inference-endpoint alignment.
        Results are success rates (\%).
        Bold denotes the highest result.
    }
    \label{tab:long_horizon_ablation}
    \vspace{1mm}

    \begingroup
    \fontsize{10pt}{12pt}\selectfont
    \setlength{\tabcolsep}{1pt}
    \renewcommand{\arraystretch}{1.1}

    \begin{tabular*}{\columnwidth}
        {@{\hspace{28pt}\extracolsep{\fill}}lcccc@{\hspace{28pt}}}
        \toprule
        \textbf{Metric}
        & \textbf{Full}
        & \textbf{w/o MC}
        & \textbf{w/o AC}
        & \textbf{w/o EA} \\
        \midrule

        $H$
        & 10 & 10 & 10 & 10 \\

        Exec.
        & 5 & 5 & 5 & 5 \\

        NFE
        & 1 & 1 & 1 & 1 \\

        \midrule

        Success (\%)
        & $\mathbf{96.0}$
        & $93.0$
        & $94.0$
        & $93.0$ \\

        \bottomrule
    \end{tabular*}
    \endgroup
\end{table}

\noindent\textbf{Results.}
Table~\ref{tab:long_horizon_ablation} reports
the component ablations on LIBERO-LONG.
The complete SOAR achieves $96.0\%$ success
with $H=10$ and five executed actions.
Removing MC, AC, or EA reduces the success
rate to $93.0\%$, $94.0\%$, and $93.0\%$,
respectively, corresponding to decreases
of 3.0, 2.0, and 3.0 percentage points.

Together with the component ablations in
Table~\ref{tab:ablation_adroit}, these results
further clarify the roles of structural
conditioning, action calibration, and
inference-endpoint alignment.
The structure head predicts observation-conditioned
temporal information, which MC incorporates into
action generation and AC uses to calibrate
the generated trajectories.

The main-text ablations show that utilizing
the predicted temporal structure through MC
and AC improves action generation.
The LIBERO-LONG results further confirm that
removing MC or AC reduces success by
3.0 and 2.0 percentage points, respectively,
while removing EA leads to a 3.0
percentage-point decrease.

Overall, these results indicate that
the benefits of structural conditioning,
action calibration, and endpoint alignment
are not confined to short-horizon 3D
manipulation. Their contributions remain
observable after integration into a VLA policy
with a longer prediction horizon.

\subsubsection{Multi-Seed Evaluation on LIBERO}
\label{app:libero_multiseed}

\noindent\textbf{Evaluation Setup.}
To assess the robustness of the VLA results, we conduct
additional evaluations on LIBERO using three independent
training seeds for $\pi_{0.5}$ with different NFEs and
$\pi_{0.5}$-SOAR.

Each evaluation covers four standard LIBERO suites,
with 50 episodes per task and 500 episodes per suite.
We report the mean and sample standard deviation
of success rates across training seeds.
The overall success rate is computed by averaging
the four suite-level results for each seed.

\begin{table}[tbp]
    \centering
    \caption{
        Multi-seed evaluation on LIBERO.
        Results are success rates (\%) reported as
        mean $\pm$ standard deviation across
        three independent training seeds.
        \textsuperscript{*} denotes baselines
        reproduced in this work.
        NFE denotes the number of action-expert
        evaluations per policy inference.
    }
    \label{tab:libero_multiseed}
    \vspace{1mm}

    \setlength{\tabcolsep}{10pt}
    \renewcommand{\arraystretch}{1.10}

    \resizebox{\textwidth}{!}{%
    \begin{tabular}{lcccccc}
        \toprule
        \textbf{Methods}
        & \textbf{NFE}
        & \textbf{Spatial}
        & \textbf{Object}
        & \textbf{Goal}
        & \textbf{Long}
        & \textbf{Average} \\
        \midrule

        $\pi_{0.5}$\textsuperscript{*}
        & 10
        & $98.60 \pm 0.35$
        & $97.87 \pm 0.31$
        & $98.00 \pm 0.20$
        & $91.53 \pm 0.81$
        & $96.50 \pm 0.35$ \\

        $\pi_{0.5}$\textsuperscript{*}
        & 2
        & $98.47 \pm 0.31$
        & $97.60 \pm 0.20$
        & $97.40 \pm 1.06$
        & $90.93 \pm 0.42$
        & $96.10 \pm 0.43$ \\

        $\pi_{0.5}$\textsuperscript{*}
        & 1
        & $97.90 \pm 0.99$
        & $96.00 \pm 3.39$
        & $97.70 \pm 0.42$
        & $90.30 \pm 0.14$
        & $95.48 \pm 0.95$ \\

        \midrule

        \textbf{$\pi_{0.5}$-SOAR}
        & \textbf{1}
        & $\mathbf{99.80 \pm 0.20}$
        & $\mathbf{97.43 \pm 0.45}$
        & $\mathbf{98.47 \pm 0.15}$
        & $\mathbf{93.13 \pm 0.64}$
        & $\mathbf{97.21 \pm 0.34}$ \\

        \bottomrule
    \end{tabular}%
    }
\end{table}

\noindent\textbf{Results.}
As shown in Table~\ref{tab:libero_multiseed},
$\pi_{0.5}$-SOAR achieves an average success rate
of $97.21\pm0.34\%$ with $\mathrm{NFE}=1$.
The reproduced $\pi_{0.5}$ baselines achieve
$96.50\pm0.35\%$ , $96.10\pm0.43\%$ and $95.48\pm0.95\%$
with $\mathrm{NFE}=10$, $2$ and $1$, respectively.

SOAR therefore improves the average success rate
by 0.71, 1.11 and 1.73 percentage points over the
reproduced 10-step ,two-step and one-step baselines,
while requiring only one action-expert evaluation.

These multi-seed results provide additional evidence
that SOAR maintains competitive task performance
across independent training runs under the
one-step inference setting.

}

\end{document}